\documentclass{article}

\usepackage{microtype}
\usepackage{graphicx}
\usepackage{booktabs} 

\PassOptionsToPackage{hyphens}{url}\usepackage{hyperref}

\usepackage[accepted]{icml2024}

\usepackage{amsmath}
\usepackage{amssymb}
\usepackage{mathtools}
\usepackage{amsthm}
\usepackage{soul}
\usepackage{enumitem}

\usepackage[capitalize,noabbrev]{cleveref}

\theoremstyle{plain}

\theoremstyle{definition}

\theoremstyle{remark}

\usepackage[textsize=tiny]{todonotes}

\usepackage{svg}
\usepackage{makecell}
\usepackage{tablefootnote}
\usepackage{multirow}
\usepackage{caption}
\usepackage{subcaption}

\definecolor{czredd}{HTML}{720026}
\definecolor{czred}{HTML}{D90429}
\definecolor{czredl}{HTML}{EF9595}
\definecolor{czredl}{HTML}{F4998D}
\definecolor{czoranged}{HTML}{CC5803}
\definecolor{czorange}{HTML}{FF8600}
\definecolor{czorangel}{HTML}{FFD89C}
\definecolor{czyellowd}{HTML}{FDC43F}
\definecolor{czyellow}{HTML}{FBE134}
\definecolor{czyellowl}{HTML}{FCEFB4}
\definecolor{czgreend}{HTML}{003E1F}
\definecolor{czgreen}{HTML}{08A045}
\definecolor{czgreenl}{HTML}{B5E48C}
\definecolor{czblued}{HTML}{01497C}
\definecolor{czblue}{HTML}{0466c8}
\definecolor{czbluel}{HTML}{78C1F3}
\definecolor{czpurpled}{HTML}{4A148C}
\definecolor{czpurple}{HTML}{7B2CBF}
\definecolor{czpurplel}{HTML}{B1B2FF}
\definecolor{czpinkd}{HTML}{EA638C}
\definecolor{czpink}{HTML}{FF8FA3}
\definecolor{czpinkl}{HTML}{FFCCD5}
\definecolor{czbrownd}{HTML}{774936}
\definecolor{czbrown}{HTML}{9D6B53}
\definecolor{czbrownl}{HTML}{DEAB90}
\definecolor{czgray}{HTML}{D1D1D1}
\definecolor{czgrayl}{HTML}{EEEEEE}

\definecolor{codegreen}{rgb}{0,0.6,0}
\definecolor{codegray}{rgb}{0.5,0.5,0.5}
\definecolor{codepurple}{rgb}{0.58,0,0.82}
\definecolor{backcolour}{rgb}{0.95,0.95,0.92}

\newcommand\best[1]{\textbf{#1}}
\newcommand\good[1]{\underline{#1}}

\icmltitlerunning{LQER: Low-Rank Quantization Error Reconstruction for LLMs}

\begin{document}

\def\LRQ{LQER}
\def\LRQn{LQER}
\def\LRQa{L$^2$QER}
\def\LRQd{LQER-dsvd}
\def\LRQaText{LLQER}
\def\mxint{\texttt{MXINT}}
\def\running{\textcolor{orange}{?}}


\twocolumn[
    \icmltitle{LQER: \underline{L}ow-Rank \underline{Q}uantization \underline{E}rror \underline{R}econstruction for LLMs}



    \icmlsetsymbol{equal}{*}

    \begin{icmlauthorlist}
        \icmlauthor{Cheng Zhang}{imperial}
        \icmlauthor{Jianyi Cheng}{cambiridge}
        \icmlauthor{George A. Constantinides}{imperial}
        \icmlauthor{Yiren Zhao}{imperial}
    \end{icmlauthorlist}

    \icmlaffiliation{imperial}{Department of Electrical and Electronic Engineering, Imperial College London, London, United Kingdom}
    \icmlaffiliation{cambiridge}{Department of Computer Science and Technology, University of Cambridge, Cambridge, United Kingdom}

    \icmlcorrespondingauthor{Cheng Zhang}{cheng.zhang122@imperial.ac.uk}
    \icmlcorrespondingauthor{Jianyi Cheng}{jianyi.cheng@cl.cam.ac.uk}
    \icmlcorrespondingauthor{George A. Constantinides}{g.constantinides@imperial.ac.uk}
    \icmlcorrespondingauthor{Yiren Zhao}{a.zhao@imperial.ac.uk}

    \icmlkeywords{Machine Learning, ICML}

    \vskip 0.3in
]



\printAffiliationsAndNotice{}  

\begin{abstract}
    Post-training quantization of Large Language Models (LLMs) is challenging.
    In this work, we introduce \textbf{L}ow-rank \textbf{Q}uantization \textbf{E}rror \textbf{R}eduction (\LRQ{}), which combines quantization and low-rank approximation to recover the model capbility.
    \LRQ{} leverages an activation-induced scale matrix to drive the singular value distribution of quantization error towards a desirable distribution, which enables near-lossless W4A8 quantization on various LLMs and downstream tasks without the need for knowledge distillation, grid search, or gradient-based iterative optimization.
    Unlike existing methods, the computation pattern of \LRQ{} eliminates the need for specialized Scatter and Gather processes to collect high-precision weights from irregular memory locations.
    Our W4A8 LLMs achieve near-lossless performance on six popular downstream tasks, while using $1.36 \times$ fewer hardware resources than the leading state-of-the-art method.
    We open-sourced our framework at \href{https://github.com/ChengZhang-98/lqer}{github.com/ChengZhang-98/lqer}.
\end{abstract}

\section{Introduction}
\label{sec:introduction}

Large Language Models (LLMs) have exhibited impressive capability on various natural language processing (NLP) tasks~\cite{brown2020language}. However, the substantial model size and its associated computation costs demand considerable energy and hardware resources. For instance, deploying BLOOM-176B~\cite{workshop2022bloom} requires 16 NVIDIA A100 GPUs and consumes more than 2000 Watts of total power ~\cite{luccioni2023estimating}. Meanwhile, empirical evidence suggests that only models with a sufficiently large parameter count begin to show \textit{emergent capabilities}~\cite{hoffmann2022training}, thereby motivates the construction of even larger models. Quantization then emerges as a promising technique to enhance the accessibility of LLMs by reducing the model size and simplifying inference computation.



\begin{figure*}
    \centering
    \begin{subfigure}[b]{0.38\textwidth}
        \centering
        \includegraphics[width=.9\textwidth]{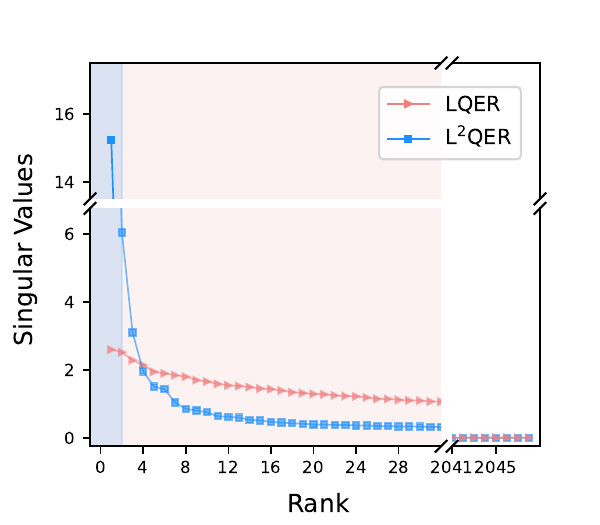}
        \caption{Singular value distributions of quantization error}
        \label{fig:s1:Eq-distribution}
    \end{subfigure}
    \hspace{0.5em}
    \begin{subfigure}[b]{0.57\textwidth}
        \centering
        \includegraphics[width=.9\textwidth]{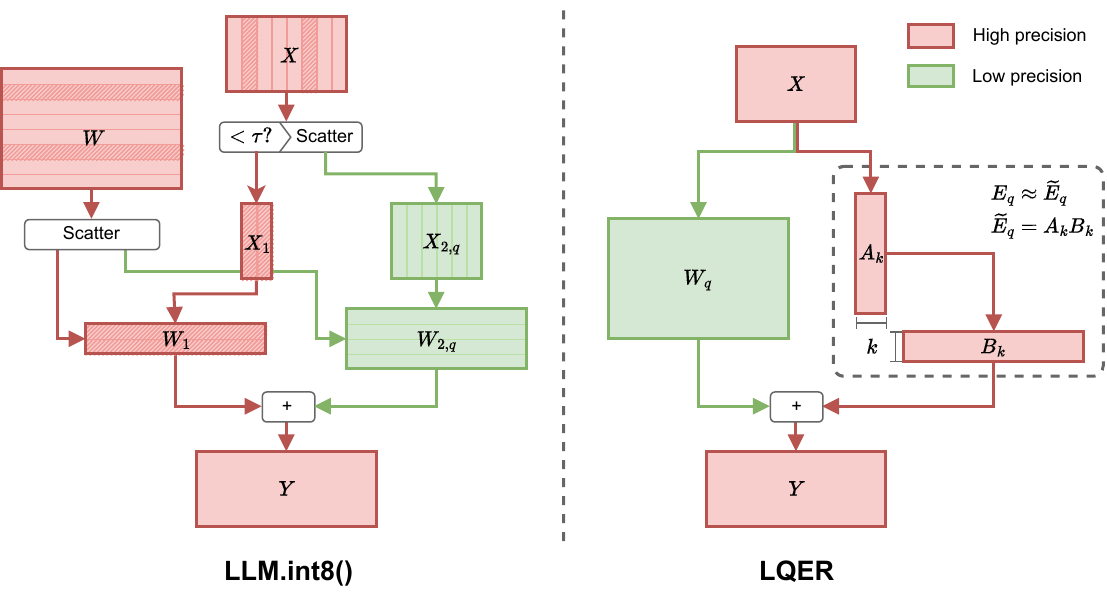}
        \caption{\texttt{LLM.int8()} v.s. \LRQ{}}
        \label{fig:s1:llm-int8-vs-lqer}
    \end{subfigure}
    \caption{Motivation and computation pattern of \LRQ{}. (a) We apply SVD to the quantization error $E_q=W-W_q$ for a 3-bit fixed-point quantized weight in OPT-1.3B, and plot their singular values distributions. Distributions are normalized to have the same Frobenius norm for a fair comparison\textsuperscript{\ref{footnote:normalization-in-fig1}}. Curves with a more asymptotic trend suggest better suitability for low-rank approximation. \LRQa{} displays a much steeper distribution with a smaller number of dominating singular values. (b) \LRQ{} approximates a trained weight $W$ with two high-precision yet low-rank matrics $A_k$ and $B_k$, and a low-precision yet high-rank matrix $W_q$. Both components are inexpensive to compute. This estbalishes a regular computation pattern that eliminates the need for irregular memory access like the Scatter and Gather operations in \texttt{LLM.int8()}.}
    \label{s1:motivation}
\end{figure*}

Low-precision Post-Training Quantization (PTQ) of LLMs has recently become an attractive solution for reducing computational and memory cost~\cite{nagel2021white}. However, it remains challenging due to the fact that 1) no further weight training is allowed and 2) the presence of magnitude outliers in model weights and activations. PTQ is a technique that quantizes a pre-trained LLM directly, without additional training, as fine-tuning LLMs usually requires substantial resources.
Many researchers have observed that
the main building block of LLMs, the transformer layer, produces magnitude outliers in weights and activations~\cite{wei2022outlier,bondarenko2021understanding,tang2023easyquant}. A simple fixed-point quantization then either suffers from considerable clipping or overflow error or from considerable rounding error, depending on the choice of scaling. In both cases, the quantization error propagates and accumulates through the LLMs, leading to substantial task accuracy degradation. To overcome this challenge, recent LLM PTQ methods investigate the statistical properties of LLMs and propose various fine-grained solutions to accommodate~\cite{dettmers2022llm,frantar2022gptq}, mitigate~\cite{xiao2023smoothquant,lee2023enhancing}, or eliminate~\cite{wei2023outlier,bondarenko2023quantizable} these numerical outliers.

However, fine-grained treatments to numerical outliers usually come with a high optimization and/or hardware cost. The optimization cost mainly stems from iterative optimization. For example, OmniQuant~\cite{shao2023omniquant} takes 7.3 hours to quantize a LLaMA-30B model with 20 iterations on a single NVIDIA A100 GPU~\cite{noauthor_omniquant_nodate}. The popular weight-only quantization setup, such as GPTQ~\cite{frantar2022gptq} and AWQ~\cite{lin2023awq}, dequantizes 4-bit weights to FP16 at runtime, which actually impedes inference on models larger than 7B~\cite{casper-hansen_autoawq}. Concurrently, many existing quantization frameworks select values from irregular positions for high-precision computation, while maintaining other values in low-precision formats~\cite{dettmers2023spqr,lee2022flexround}. For instance, \texttt{LLM.int8()}~\cite{dettmers2022llm} selects activation outliers to compute in half-precision floating-point, while casting the rest to integers. In this work, we propose a simple and efficient LLM PTQ framework that avoids iterative optimization and irregular computation patterns.


Optimizing weight quantization can be considered as a process of minimizing the quantization error $E_q = W - W_q$, where $W_q$ is the quantized weights.
We are firstly interested in a novel \textit{inference framework formulation} termed \textit{\LRQ{}}. \LRQ{} approximates the real value of $W$ through two components ($W \approx \widetilde{E}_q + W_q$): a high-precision yet low-rank matrix $\widetilde{E}_q$ that approximates $E_q$ but with $\mathrm{rank}(\widetilde{E}_q) \ll \mathrm{rank}(W)$; and a low-precision yet high-rank matrix $W_q$ as shown in \Cref{fig:s1:llm-int8-vs-lqer}. Both components  are inexpensive to compute and thus work together to reduce the overall computational complexity. Crucially, the high-precision, low-rank component $\widetilde{E}_q$ establishes a regular computation pattern that eliminates need of having the Scatter and Gather operations to fetch and store values from irregular memory locations like \texttt{LLM.int8()} (\Cref{fig:s1:llm-int8-vs-lqer}).

\begin{table*}[ht]
    \caption{A summary of recent LLM PTQ methods. Weight-only ($w$-only) and weight-activation ($w\&a$) quantizations are two popular setups.
    $w$-only quantization generally dequantize values (($\mathrm{dq}(\cdot)$)) back to FP16 before the weight-activation matrix mutiplication at inference time.
    $w\&a$ quantization performs low-precision mutiplication ($X_qW_q$) at inference-time but requires finding an invertible matrix $S$ to decrease the magnitude range of activations (detail explained in \Cref{sec:related:ptq}).
    We shortlist the recent works that belong to the two setups in the last column. $^*$ indicates the common precision of $W_q$ and $X_q$ that achieves almost lossless performance on downstream tasks.
    }
    \begin{center}
        \begin{small}
            \resizebox{\linewidth}{!}{
                \begin{tabular}{lllll}
                    \toprule
                    Q setup  & WxAy$^*$ & Quantization function                               & Inference-time                           & Methods       \\
                    \midrule
                    $w$-only & W4       & $\begin{array} {r@{}l@{}}(W_q, \mathbf{s}) =\mathrm{q}(W)\end{array}$                & $\begin{array} {r@{}l@{}}\widetilde{Y}=X\mathrm{dq}(W_q, \mathbf{s})\end{array}$ & \makecell[l]{GPTQ~\cite{frantar2022gptq}, AWQ~\cite{lin2023awq},\\Z-fold~\cite{jeon2023frustratingly}, QuiP~\cite{chee2023quip},\\FlexRound~\cite{lee2022flexround}, LRQ~\cite{luo2023long}} \\ \midrule
                    $w\&a$  & W8A8     & $\begin{array} {r@{}l@{}} &(X_q, \mathbf{s}_{t})=\mathrm{q}(XS)\\ &(W_q, \mathbf{s}_c)=\mathrm{q}(S^{-1}W)\end{array}$ &     $\begin{array} {r@{}l@{}} &\widetilde{Y}_{i,j} = \mathbf{s}_{t,i}\mathbf{s}_{c,j} (X_{q,i,:} \cdot X_{q,:,j}) \\ & (Y_{q,i,:}, \mathbf{s}'_{t,i}) = \mathrm{q}([\widetilde{Y}_{i,1}, \widetilde{Y}_{i,2}, \dots]) \end{array}$                              & \makecell[l]{SmoothQuant~\cite{xiao2023smoothquant}, OS+\cite{wei2023outlier},\\AQAS~\cite{lee2023enhancing}, OmniQuant~\cite{shao2023omniquant}}\\
                    \bottomrule
                \end{tabular}
            }
        \end{small}
    \end{center}
    \label{tab:s2:related-work}
\end{table*}


In this study, we explore optimizations for $W_q$ using both the integer format and the recently proposed MX number formats~\cite{rouhani2023microscaling}. Additionally, our work emphasizes the design of $\widetilde{E}_q$.
Theoretically, assuming the trained weights to be independent and identically distributed (i.i.d.), and given a sufficiently high chosen precision, $E_q$ can be approximated as a random matrix formed by the round-off error. The Marchenko–Pastur distribution suggests that \textit{there exihibits an asymptotic behavior for the distribution of singular values of large random matrices}~\cite{marchenko1967}. We then show the actual singular value distributions of $E_q$ in \Cref{fig:s1:Eq-distribution} from a linear layer in OPT-1.3B~\cite{zhang2022opt} (labeled as \LRQ{}), showcasing a similar phenomenon to what the \textit{Marchenko–Pastur law} has suggested. Further motivated by the fact that matrices with only a few large singular values like $E_q$ can be effectively approximated by low-rank matrices.
We propose to left-multiply $E_q$ by a diagonal matrix $S$, derived from activation magnitudes, that pushes the singular values of $E_q$ toward an even more desirable distribution (labeled as \LRQa{} in \Cref{fig:s1:Eq-distribution}).
The singular values of $SE_q$ decay more rapidly than $E_q$, with the large singular values of $SE_q$ concentrates in the first few components, as illustrated in \Cref{fig:s1:Eq-distribution}\footnote{\label{footnote:normalization-in-fig1}To make a fair comparison, we normalize $E_q$ before SVD by multiplying $E_q$ with a scalar $\alpha$ such that the scaled $\alpha E_q$ has the same Frobenius norm as $SE_q$.}. This observation then further motivates our LLM Post-Training Quantization (PTQ) method, \textit{\textbf{L}eft-multiply \LRQ{} (\LRQa{})}, designed to recover the performance loss caused by quantization.
We make the following contributions in this work:
\begin{itemize}[leftmargin=*]
    \item We introduce a novel quantized LLM inference framework termed \textbf{L}ow-rank \textbf{Q}uantization \textbf{E}rror \textbf{R}eduction (\LRQ{}) which combines quantization and low-rank approximation. Unlike existing methods that require gathering values from irregular memory locations, \LRQ{} boasts a blocked and regular computation pattern and employs a unified number format for both memory and computation.
    \item We then propose \LRQa{}, a straightforward but efficient quantization method on top of \LRQ{}. \LRQa{} does not need any expensive knowledge distillation, hyper-parameter search, or other forms of iterative optimization. We showcase \LRQa{}'s competitiveness with current state-of-the-art methods. \LRQa{} quantizes both weights and activations, it pushes the limit to W4A6, matching the perplexity of OmniQuant (W6A6) on WikiText. Compared to weight-only ($w$-only) quantization methods, our approach outperforms AWQ (W4A16) and maintains quantization activations staying at 8-bit (W4A8).

\end{itemize}

\section{Related Work}

\subsection{Post-Training Quantization of LLMs}
\label{sec:related:ptq}
Post training quantization of LLMs is a challenging task due to presence of numerical outliers. 
Existing methods can be broadly categorized into two setups: weight-only ($w$-only) and weight-activation ($w\&a$) quantizations. Recent works within these two setups are summarized in~\Cref{tab:s2:related-work}.

\paragraph{Weight-only quantization} Weight-only quantization usually partitions the trained weight matrix $W$ into groups, with the $i$-th group being quantized using a scale factor $\mathbf{s}_i$:
\begin{equation}
    (W_q, \mathbf{s}) = \mathrm{q}(W)
    \label{eq:s2:w-only-quantization}
\end{equation} 
where $W_q$ is the quantized weight matrix, $\mathbf{s}$ is a vector of scale factors, and $\mathrm{q}(\cdot)$ denotes quantization function.
During inference, the low-precision weights $W_q$ is dequantized back to FP16 before the weight-activation matrix multiply:
\begin{equation}
    \widetilde{Y} = X\mathrm{dq}(W_q, \mathbf{s})
    \label{eq:s2:w-only-quantization-inference}
\end{equation}
Here $X$ is the FP16 input, and $\mathrm{dq}(\cdot)$ is the dequantization function, and $\widetilde{Y}$ is the output. The runtime dequantization cost is negligible in memory-bound scenarios, \textit{e.g.}, small models at small batch sizes. \textit{This cost escalates with model sizes, and eventually impedes inference in compute-bound scenarios}~\cite{casper-hansen_autoawq}.

GPTQ~\cite{frantar2022gptq} and AWQ~\cite{lin2023awq} are two representative $w$-only quantization methods. GPTQ employs second-order information to iteratively round grouped weights and correct the quantization error in the remaining groups.
AWQ protects salient weights induced by activations using per-channel scaling. Recent advancements in $w$-only setup include Z-Fold~\cite{jeon2023frustratingly} and QuiP~\cite{chee2023quip}, following GPTQ to correct quantization error. FlexRound~\cite{lee2022flexround}, and LRQ~\cite{luo2023long} follow AWQ to study finer-grained weight scaling.


\paragraph{Weight-activation quantization} $w\&a$ quantization utilizes an invertible matrix $S$ to reduce the magnitude range of activations before quantization:
\begin{equation}
    (X_q, \mathbf{s}_t) = \mathrm{q}(XS)
    \label{eq:s2:w-a-quantization}
\end{equation}
where $\mathbf{s}_t$ is a vector a per-token scalars. $S^{-1}$ is fused into the weight matrix $W$, and $S$ is fused into the weight matrix of the preceding layer before quantization:
\begin{equation}
    (W_q, \mathbf{s}_c) = \mathrm{q}(S^{-1}W)
    \label{eq:s2:w-a-quantization-inference}
\end{equation}
where $\mathbf{s}_c$ is a vector of per-channel scalars. At inference time, the inner product in the activation weight matrix multiplication consists of an inner product of two fixed-point vectors and an FP16 multiplication between the token scalar and channel scalar (``Inference-time'' entry in~\Cref{tab:s2:related-work}). Then each row of output activation matrix is quantized back to the input format. This style of quantization avoids the extra dequantization cost in $w$-only setup, but achieving a precision lower than W8A8 while maintaining nearly-lossless model capability proves challenging. Existing $w\&a$ quantization methods lower than 8-bit precision usually suffer from an average downstream task accuracy drop larger than 1\%~\cite{shao2023omniquant,liu2023qllm}.

SmoothQuant~\cite{xiao2023smoothquant} pioneered fusion of an invertible scale matrix into its preceding layer. Outlier Suppression+~\cite{wei2023outlier} further introduces a bias matrix to~\Cref{eq:s2:w-a-quantization-inference} to shift the mean of activations towards zero and update the layer bias accordingly. Recent works following this line of research include AQAS~\cite{lee2023enhancing}, and OmniQuant~\cite{shao2023omniquant}.

Another unique $w\&a$ quantization method, \texttt{LLM.int8()} decomposes the FP16 matrix multiplication into a 8-bit fixed-point and an FP16 sub-matrix multiplication using activation thresholds. Despite achieving the closest model capability to FP16, the thresholding, Scatter and Gather operations of \texttt{LLM.int8()} are expensive in large models. Similar to \texttt{LLM.int8()}, SpQR~\cite{dettmers2023spqr} and EasyQuant~\cite{tang2023easyquant} are recent works that retains salient weights in FP16 at finer granularity while quantizing the rest to low-precision.

In this work, we propose a fundamentally different PTQ framework that approximates the real value of weight through two components ($W = \widetilde{E_q} + W_q$), and demonstrate how this formuation helps us to achieve almost lossless PTQ in the $w\&a$ quantization setup with a $W4A8$ configuration.


\subsection{The \mxint{} Arithmetic}

Block floating point is a family of number formats that represents a vector of numbers using a shared exponent or exponent bias. Various block floating point formats have been explored for efficient training or inference in the past few years~\cite{darvish2020pushing,fox2020block,zhang2022fast,drumond2018training}. One notable representative is \mxint{}, introduced for hardware-efficient post training quantization~\cite{darvish2020pushing, rouhani2023shared}, this number format has recently been standardized by AMD, Arm, Intel, Meta, Microsoft, NVIDIA, and Qualcomm, for next-generation AI facilities~\cite{mxspecs2023}.

\begin{figure}[t]
    \begin{center}
        \centerline{\includegraphics[width=0.45\textwidth]{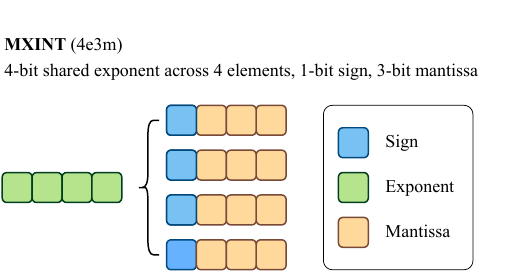}}
    \end{center}
    \caption{\mxint{} number format ~\cite{rouhani2023microscaling}. \mxint{} places a shared exponent across a group of fixed-point numbers.
        \mxint{} is more hardware efficient than floating point for its simplified vector inner product, and provides a large dynamic range compared to fixed-point numbers. \mxint{} has been standardized recently for next generation AI hardware systems~\cite{mxspecs2023}.
    }

    \label{fig:s2:mxint}

\end{figure}

\Cref{fig:s2:mxint} illustrates an example of an \mxint{} vector sharing a 4-bit exponent across four 4-bit mantissas. \mxint{} excels in hardware efficiency compared to floating point, as the inner product of two \mxint{} vectors can be computed as a inner product of two fixed-point numbers plus an exponent addition. Meanwhile, the shared exponent provides a larger dynamic range than fixed point numbers. Recent works indicate that this extended dynamic range fits the activation outliers well in LLM PTQ tasks~\cite{zhang-etal-2023-revisiting,rouhani2023microscaling}. In this work, We adopt \mxint{} as the default number format while the idea can be applied to other formats. 

\subsection{Low-Rank Adapters for Fine-Tuning}

Low-rank adapter (LoRA)~\cite{hu2021lora} is a parameter efficient fine-tuning method for saving GPU memory. LoRA freezes the pretrained weight $W$, and only updates two low-rank weight matrices $L_1$ and $L_2$ during fine-tuning. Based on LoRA, QLoRA~\cite{dettmers2023qlora} keeps the quantized pretrained weights in memory and only double-dequantizes\footnote{Double-dequantize means the dequantization scales the 4-bit weight matrix twice using $c_1^{\text{FP32}}$ and $c_2^{\text{k-bit}}$} it in the forward pass to further reduce fine-tuning memory footprints:
\begin{equation}
    \begin{split}
        Y^{\text{BF16}} = X^{\text{BF16}} \mathrm{ddq}(c_1^{\text{FP32}},c_2^{\text{k-bit}}, W^{\text{NF4}}) \\+ X^{\text{BF16}}L_1^{\text{BF16}}L_2^{\text{BF16}}
    \end{split}
\end{equation}
The advantage of LoRA-based methods is that the fine-tuned model can be deployed without extra cost as the low-rank matrices are fused into the pretrained weights after fine-tuning . For QLoRA, the fusion can be expressed as:
\begin{equation}
    W^{\text{BF16}}_{\text{new}} = \mathrm{ddq}(c_1^{\text{FP32}},c_2^{\text{k-bit}}, W^{\text{NF4}}) + L_1^{\text{BF16}}L_2^{\text{BF16}}
\end{equation}
LoftQ~\cite{li2023loftq} initializes $L_1$ and $L_2$ with the Singular Value Decompostion (SVD) of quantization errors to achieves a faster fine-tuning convergence than QLoRA.

To our knowledge, LoftQ is the closest work to ours. However, our \LRQ{} framework is fundamentally different from the above as it is a PTQ method that does not target fine-tuning. The core idea of \LRQ{} is that shaping the singular value distribution of quantization error approximator ($\widetilde{E_q}$) enables a nearly-lossless inference pass quantization. LoftQ fuses the low-rank matrices back to original FP32 weights when deployed, however, the low-rank matrices in \LRQ{} remains separate from the quantized weight matrix $W_q$: this allows the matrix multiplications for the low-precision high-rank weight matrix ($W_q$) and the low-rank high-precision matrices to happen in parallel at inference time.

\section{Method}

We aim to approximate the multiplication by a large dense weight matrix $W$ in a low-cost way. This low cost can be achieved through low-precison quantization or low-rank approximation.
Quantization simplifies the multiplication arithmetic, while low-rank approximation reduces the overall number of computations. We judiciously combine the two: approximate $W$ as a dense low-precision matrix $W_q$ and then correct the error induced using a high-precision low-rank correction term as illustrated in \Cref{fig:s1:llm-int8-vs-lqer}.

\subsection{\LRQn: Approximate \texorpdfstring{$E_q$}{Eq} using SVD}
\begin{figure}
    \vskip 0.2in
    \begin{center}
        \includegraphics[width=0.45\textwidth]{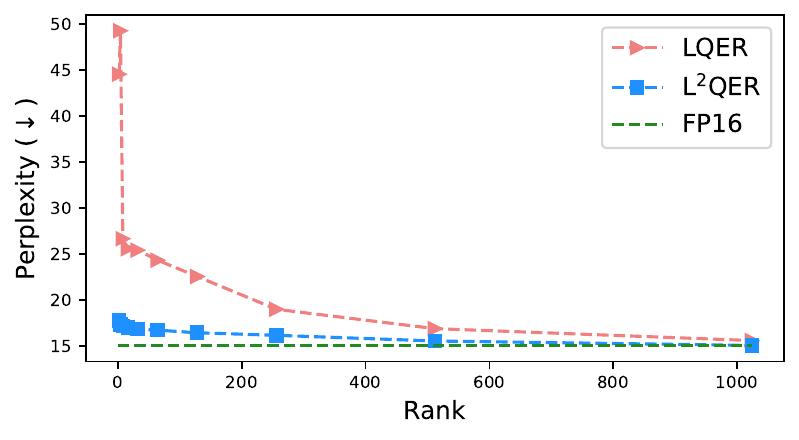}
        \vskip -0.1in
        \caption{ Perplexity ($\downarrow$) vs rank. We apply W3A8 \LRQn{} and \LRQa{} to OPT-1.3B and plot the resultant perplexity. Considering the embedding dimension is 2048, \LRQn{} requires a fairly large $k\approx 600$ to reach a perplexity close to FP16 . In comparison, a small $k\approx 64$ is enough for \LRQa{} Comparison of perplexity ($\downarrow$) and quantization error reconstruction between \LRQn{} and \LRQa{}. 
        }
        \label{fig:s3:method-naive-vs-act}
    \end{center}
    \vskip -0.2in
\end{figure}


Our idea is to reconstruct the quantization error matrix $E_q$ through SVD-based low rank approximation. When a quantization is applied to a trained FP32/FP16 weight matrix $W \in \mathbb{R}^{m \times n}$, the resulting quantization error matrix $E_q$ is:
\begin{equation}
    E_q = W - W_q
\end{equation}
where $W_q=\mathrm{q}(W)$ is the quantized weight matrix, and $\mathrm{q}(\cdot)$ represents the quantization function. A straightforward way to reconstruct the error is to use SVD-based low-rank approximation:
\begin{equation}
    E_q = U \Sigma V^T \approx U_k \Sigma_k V_k^T
\end{equation}
where $U \in \mathbb{R}^{m \times m}$ and $V \in \mathbb{R}^{n \times n}$ are orthogonal matrices, $\Sigma \in \mathbb{R}^{m \times n}$ is a diagonal matrix of singular values. $U_k \in \mathbb{R}^{m \times k}$, $V_k \in \mathbb{R}^{n \times k}$ and $\Sigma_k \in \mathbb{R}^{k \times k}$ are the sub-matrices of $U$, $V$ and $\Sigma$ corresponding to the largest $k$ singular values.

If two high-precision matrices $A_k = U_k$ and $B_k = \Sigma_k V_k^T$ are assigned to approximate $E_q$, \textit{i.e.}, $A_k B_k \approx E_q$, the linear layer can be approximated as:
\begin{equation}
    \begin{split}
        \widetilde{Y} & = X W_q + (X A_k) B_k       \\
                      & = X (W_q + A_k B_k)         \\
                      & = X (W_q + \widetilde{E_q}) \\
                      & \approx X (W_q + E_q)       \\
                      & = X W
    \end{split}
    \label{eq:lrhp-naive-layer}
\end{equation}
where $X\in \mathbb{R}^{t\times m}$ and $\widetilde{Y}\in \mathbb{R}^{t\times n}$ are the layer input and the approximated output, and $t$ is the sequence length. We use $b_l$ and $b_h$ to denote the bitwidth of low-precision matrix ($W_q$) and high-precision matrices ($X$, $A_k$ and $B_k$) respectively. A pair of $b_l$ and $b_h$, \textit{e.g.}, $(b_l, b_h)=(3, 8)$, means that we compensate the quantization error of 3-bit weight using two 8-bit low-rank matrices. We refer to this design of the inference flow as \LRQn{}.

At inference-time, \LRQ{} runs one low-precision but large matrix multiplication ($XW_q$) and two high-precision but small matrix multiplications ($XA_k$ and $(XA_k)B_k$) in parallel to save memory and achieve a speedup. Given a low-precision quantization $\mathrm{q}(\cdot)$, adjusting the rank $k$ allows tuning the trade-off between the computational cost and the model accuracy. Specifically:
\vspace{-3pt}
\begin{itemize}[leftmargin=*]
    \item In LLMs, $W \in \mathbb{R}^{m\times n}$ is usually a large matrix. For example, $(m, n)$ is $(12288, 12288)$ , $(12288, 49152)$, or $(49152, 12288)$ in OPT-175B. A low-precision $W_q$ significantly reduces the memory footprint and $X W_q$ is faster than $XW$.  
    \item Two high-precision but small matrices $A_k \in \mathbb{R}^{m \times k}$ and $B_k \in \mathbb{R}^{k \times n}$ estimate the quantization error at the cost of minimal computation. For a token $\mathbf{x}\in \mathbb{R}^{m}$, the two matrix multiplies $(\mathbf{x} A_k)$ and $((\mathbf{x} A_k) B_k)$ only introduce $(m+n) \times k$ high-precision multiplies in total while the unquantized $\mathbf{x} W$ has $m \times n$ high-precision multiplies. For the FNN layers in OPT-175B, the newly introduced multiplications is around $\frac{(m+n)\times k}{m \times n} \approx 0.01 \times k \%$.
\end{itemize}
\vspace{-3pt}
The ideal case is that a small $k \ll \min(m, n)$, \textit{e.g.}, $k=32$, would successfully recover the model's accuracy/perplexity. However, our experiments reveal that the singular values of $E_q$ decay slowly for most linear layers, requiring a sufficiently large $k$ to recover the accuracy/perplexity. ~\Cref{fig:s3:method-naive-vs-act} illustrates the perplexity of quantized W3A8 OPT-1.3B versus rank $k$. For \LRQn{}, a small $k \approx 64$ still falls short compared to the FP16 baseline. The following section then discusses how we can achieve a low $k$ value by analytically scaling the error term.

\subsection{\texorpdfstring{\LRQa{}}{\LRQaText{}}: Shape Singular Value Distribution of Quantization Errors using Activation Statistics}
\label{sec:method:lrqa}

Recent works have shown that partially preserving the weight precision according to activation magnitude recovers the model's accuracy/perplexity. \texttt{LLM.int8()} decomposes an FP16 matrix multiply into one FP16 sub-matrix multiply for large activation magnitudes and one 8-bit fixed-point sub-matrix multiply for the rest at runtime. AWQ also presents an experiment that effectively recovers accuracy by preserving the 1\% salient weights corresponding to large activation magnitudes in FP16, and quantizing other weights to 4-bit grouped fixed-point.

Motivated by this phenomenon, we propose a novel quantization error reconstruction method, named \LRQa{}, that scales the quantization error matrix $E_q$ before applying SVD and undo the scaling in low-rank matrices. We first left-multiply $E_q$ with a diagonal matrix $S=\text{diag}(s_1, s_2, \dots, s_m)$ to scale $i$-th row of $E_q$ by a distinct scalar $s_i$, then apply SVD to the scaled matrix $S E_q$:
\begin{equation}
    S E_q = U' \Sigma' V'^T \approx U'_k \Sigma'_k {V'}_k^T
\end{equation}
where $S$ is calibrated from the pre-training data. To calculate $s_i$, we first average the $i$-th channel magnitudes across all the tokens in each calibration sample, then find the maximum average value among all samples. A detailed calculation of $S$ is in~\Cref{sec:appendix:data-calibration}. The calibration requires no training.

The intuition behind the scaling is that the quantization error corresponding to large activation magnitudes, \textit{i.e.}, the salient weights identified by corresponding activation magnitudes, should be more precisely approximated. Hence, we scale up these quantization errors before SVD.

High precision $A'_k$ and $B'_k$ are employed to cancel out $S$ and reconstruct $E_q$:
\begin{equation}
    \left\{
    \begin{aligned}
        A'_k & = S^{-1} U'_k        \\
        B'_k & = \Sigma'_k {V'}_k^T
    \end{aligned}
    \right.
\end{equation}
where $S^{-1}$ is the inverse of the diagonal matrix $S$. $S^{-1}$ always exists in practice since no diagonal elements in $S$ is zero (no channels in LLM activations are always zero). Now we approximate the linear layer similarly to \LRQn:
\begin{equation}
    \begin{split}
        \widetilde{Y} & = XW_q + (X A'_k) B'_k                             \\
                      & = XW_q + (X S^{-1} U'_k) (\Sigma'_k {V'}_k^T)      \\
                      & = X \left(W_q + S^{-1} \widetilde{(SE_q)}_k\right) \\
                      & \approx XW
    \end{split}
\end{equation}
where $\widetilde{Y}$ and $\widetilde{(SE_q)}_k=U'_k \Sigma'_k {V'}_k^T$ are the approximated $Y$ and the approximated $(SE_q)$ of rank $k$. Note that the term $\widetilde{E}_q:=S^{-1} \widetilde{(SE_q)}_k$ is the approximated quantization error.


As shown in~\Cref{fig:s1:Eq-distribution}, $S$ drives the singular value distribution to decay faster than \LRQn{} with large singular values concentrating in the first few components, and the scaling is counteracted by $S^{-1}$ in $A'_k$; therefore, \LRQa{} tends to recover more model capability than \LRQn{}. In~\Cref{fig:s3:method-naive-vs-act}, \LRQa{} recovers the perplexity close to FP16 baseline at a very small $k\approx 64$. In~\Cref{sec:experiments:w4a8-quantization}, we will show that \LRQa{} achieves nearly lossless W4A6 LLM PTQ results comparable to state-of-the-art W6A6/W4A16 methods but with higher hardware efficiency.

\section{Experiments}
\label{sec:experiments}

\begin{table}[t]
    \caption{Perplexity ($\downarrow$) of plain \mxint{}, \LRQn{}, and \LRQa{} on OPT-1.3B and LLaMA-7B. We apply plain \mxint{} quantization, \LRQn, and \LRQa{} to OPT-1.3B and LLaMA-7B in the same W4A8 setup. The decreasing perplexity proves the effectiveness of the quantization error reconstruction in \LRQn{}, and activation-induced scale matrix $S$ in \LRQa{}.}
    \begin{center}
        \begin{small}
            \begin{tabular}{lcccc}
                \toprule
                              & \mxint{}    & \LRQn{}    & \LRQa{}    & FP16  \\ \midrule
                OPT-1.3B           & 16.42    & 15.28   & 15.02   & 14.63 \\
                $\Delta$ PPL ($\downarrow$) & +1.78       & +0.65      & +0.39      & -     \\  \midrule
                LLaMA-7B           & 6.17      & 6.06    & 5.89    & 5.67 \\
                $\Delta$ PPL ($\downarrow$)   & +0.50       & +0.39      & +0.22      & - \\ \bottomrule
            \end{tabular}
        \end{small}
    \end{center}
    \vskip -0.1in
    \label{tab:s4:ablation-study-4-cases}
\end{table} 

\subsection{Experimental Setup}

\paragraph{Quantization} We use \mxint{} as the number format of \LRQ{} if not specified. In~\Cref{sec:experiments:w4a8-quantization}, we use W4A8 \LRQa{} with $k=32$ to compare with both 4-bit $w$-only and 4-/6-/8-bit $w\&a$ quantization methods. In~\Cref{sec:experiments:2bit-quantization}, we use W2A8 \LRQa{} with $k=256$ to compare with 2-bit $w$-only quantization methods. In both subsections, \mxint{} activation matrices have 8-bit shared exponents to accomodate activation outliers, while weight matrices and low-rank matrices have 4-bit shared exponents. The block size of \mxint{} is the default [1, 16] in the original paper~\cite{darvish2020pushing} for $X_q$ ([16, 1] for $W_q$, $A_k$, and $B_k$).

\paragraph{Models and Baselines} We benchmarked our methods on the OPT family~\cite{zhang2022opt}, the LLaMA family (including LLaMA~\cite{touvron2023llama}, LLaMA-2~\cite{touvron2023llama2}, Vicuna-v1.5~\cite{zheng2023judging}), and Mistral~\cite{jiang2023mistral}. These are the representative or state-of-the-art model open-sourced for research across various model sizes and architectures.
\begin{table*}[ht]
  \caption{A comparison of perplexity($\downarrow$) on WikiText-2. Best results are marked in \best{bold}, and second best results are \good{underlined} in $w\&a$ setup.
    In a $w$-only setup, \LRQa{}-\texttt{INT} outperforms GPTQ and is on par with AWQ, while offering substantially lower hardware costs.
    In a $w\&a$ setup, \LRQa{}-\mxint{} outperforms all other competitors both in terms of perplexity and hardware efficiency.
    $^*$ means \texttt{LLM.int4()} casts the weight sub-matrices corresponding to activation outliers to 4-bit fixed-point before computation and cast them back to FP16 after, thus the weight formats in memory is FP16. $\dag$ means OmniQuant and AQAS use per-channel and per-token scaled quantization. $^\ddag$ means LLaMA-2 results were not available in~\cite{lee2023enhancing} and the author has not open-sourced AQAS code.}
  \begin{center}
    \begin{small}
      \resizebox{\linewidth}{!}{
        \begin{tabular}{lllcccccccccccc}
          \toprule
          \multirow{2}{*}{Q Setup}    &
          \multirow{2}{*}{Method}     &
          \multirow{2}{*}{Q Config}   &

          \multicolumn{3}{c}{OPT}     &
          \multicolumn{3}{c}{LLaMA}   &
          \multicolumn{3}{c}{LLaMA-2} &
          \multirow{2}{*}{\makecell{Avg.                                                                                                                                                                                                                      \\$\Delta$ PPL ($\downarrow$)}} &
          \multirow{2}{*}{\makecell{Avg.                                                                                                                                                                                                                      \\$w$ bits}}          &
          \multirow{2}{*}{\makecell{Circuit area                                                                                                                                                                                                              \\($\downarrow$)}}                                                                                                                                             \\
          \cmidrule(lr){4-6} \cmidrule(lr){7-9} \cmidrule(lr){10-12}

                                      &                      &               & 6.7B         & 13B          & 30B         & 7B          & 13B         & 33B         & 7B          & 13B         & 70B         &             &            &                     \\ \midrule
          -                           & FP16                 & -             & 10.86        & 10.13        & 9.56        & 5.67        & 5.10        & 4.10        & 5.48        & 4.90        & 3.32        & -           & 16         & 1$\times$           \\ \midrule
          \multirow{3}{*}{$w$-only}   & GPTQ                 & INT4, g128    & 11.39        & 10.31        & 9.63        & 6.12        & 5.21        & 4.24        & 5.69        & 4.98        & 3.51        & 0.22        & 4.1        & 13.99$\times$       \\
                                      & AWQ                  & INT4, g128    & \best{10.93} & \best{10.21} & 9.59        & \best{5.78} & \best{5.20} & \best{4.22} & 5.61        & 4.98        & \best{3.42} & \best{0.09} & 4.1        & 13.99$\times$       \\
                                      & \LRQa{}-\texttt{INT} & INT4, g128    & 10.99        & 10.24        & \best{9.57} & 5.89        & \best{5.20} & 4.24        & \best{5.58} & \best{4.96} & \best{3.42} & 0.11        & 4.3        & \best{1.34}$\times$ \\ \midrule
          \multirow{6}{*}{$w\&a$}     & \texttt{LLM.int4()}  & $\tau=6.0$    & 11.23        & 10.39        & 10.01       & 6.05        & 5.31        & 4.33        & 5.77        & 5.06        & 3.51        & 0.29        & 16$^*$     & $ 21.23\times$      \\
                                      & OmniQuant $\dag$     & W6A6, per-c/t & \best{10.96} & \best{10.21} & \best{9.62} & 5.96        & 5.28        & 4.38        & 5.87        & 5.14        & 3.72        & 0.20        & 6.0        & $ 0.39\times$       \\
                                      & AQAS  $\dag$         & W4A8, per-c/t & 13.42        & 12.19        & 11.08       & 6.69        & 5.81        & 5.14        & -$^\ddag$   & -$^\ddag$   & -$^\ddag$   & 1.45        & \best{4.0} & 0.45$\times$        \\
                                      & \LRQa{}-\texttt{INT} & W4A8, g128    & 11.10        & 10.38        & 9.72        & 6.09        & 5.31        & 4.35        & 5.85        & 5.10        & 3.51        & 0.25        & \good{4.1} & \good{0.33$\times$} \\
                                      & \LRQa{}-\mxint{}     & W4A6          & 11.03        & 10.32        & 9.72        & \good{5.92} & \good{5.24} & \good{4.28} & \good{5.73} & \good{5.05} & \good{3.46} & \good{0.18} & 4.3        & \best{0.23$\times$} \\
                                      & \LRQa{}-\mxint{}     & W4A8          & \good{11.00} & \good{10.27} & \good{9.69} & \best{5.89} & \best{5.21} & \best{4.25} & \best{5.69} & \best{5.02} & \best{3.44} & \best{0.15} & 4.3        & \good{0.33$\times$} \\ \bottomrule
        \end{tabular}
      }
    \end{small}
  \end{center}
  \vskip -0.15in
  \label{tab:s4:w4a8-perplexity}
\end{table*}
We compare our methods with FP16 model, \texttt{LLM.int4()}\footnote{\texttt{LLM.int4()} denotes the 4-bit verision of \texttt{LLM.int8()} open-sourced in \hyperlink{https://github.com/TimDettmers/bitsandbytes}{bitsandbytes}.}, GPTQ, AWQ, AQAS, OmniQuant\footnote{We take W6A6 OmniQuant as an weight-activation quantization baseline, and W2A16 as a 2-bit weight-only quantization baseline.}, and QuiP. The later two have variants optimized for extremely low-precision quantization. We take the reported WikiText2 perplexity or downstream task accuracy from the original papers if available.

\paragraph{Evaluation} We report the perplexity on WikiText-2~\cite{merity2016pointer} and the accuracy on ARC (easy)~\cite{allenai_arc}, ARC (challenge)~\cite{allenai_arc}, LAMBADA~\cite{paperno2016lambada}, PIQA~\cite{Bisk2020}, OpenBookQA~\cite{OpenBookQA2018}, and BoolQ~\cite{clark2019boolq} using the \texttt{lm-eval-harness} evaluation flow~\cite{eval-harness}. Ideally a calibration dataset should be sampled from the pretraining dataset to calculate the activation-induced scale matrix $S$. However, none of the LLMs mentioned above open-sourced their pretraining datasets. We create a subset of SlimPajama~\cite{cerebras2023slimpajama} with Wikipedia texts excluded as the calibration dataset. This calibration dataset contains only 32 samples of 2048 tokens.
As mentioned previously in \Cref{sec:method:lrqa}, this calibration simply profiles values without having any SGD-based training.
We also report the weight average bitwidth for memory efficiency and estimate the circuit area for the hardware cost. Circuit area is estimated with the number of Look Up Tables (LUTs) of the processing engines (PEs) if implemented on FPGAs, which is also approximately proportional to the number of gates if implemented as ASICs. We have faithfully implemented these arithmetic cores and inputted them into FPGA synthesis flows, obtaining results for circuit area. This is because \mxint{} is a newly release arithmetic standard~\cite{mxspecs2023}. \Cref{sec:appendix:hardware-cost} provides the detailed circuit area estimation.

\subsection{\LRQ{} and \LRQa{}}

We first focus on comparing variants of \LRQ{} in \Cref{tab:s4:ablation-study-4-cases}.
We evaluate the variants in a W4A8 $w\&a$ quantization setup on both OPT-1.3B and LLaMA-7B.
We show the results of plain \mxint{}, \LRQn{}, and \LRQa{}, where plain \mxint{} means the whole network is simply \mxint{} quantized without any special treatments.

\Cref{tab:s4:ablation-study-4-cases} indicates that a plain W4A8 \mxint{} quantization leads to substantial performance degradation ($\Delta \text{PPL}$ = +1.78 on OPT-1.3B).
\LRQn{} verifies that reconstructing the quantization error of weight helps to recover the model performance. Activation-induced $S$ in \LRQa{} further pushes the performance of \LRQn{} to be even closer to the FP16 baseline. In the following sections, we then mainly focus on presenting \LRQa{} results.

\subsection{Comparing with Existing Quantization Methods}
\label{sec:experiments:w4a8-quantization}

\begin{table*}[t]
    \caption{A comparison of downstream task accuracy ($\uparrow$), averaged across six downstream tasks. \best{Bold} text indicates the best results, while \good{underscore} denotes the second-best. \LRQa{} \textit{achieves the best accuracy among all LLaMA models}, and nearly lossless (around 0.3\% drop) compared to the FP16 baseline. $^{*}$ means the results are not available in the original GPTQ paper, and we did not find open-source implementations and/or model checkpoints to run evaluation. $^{\dag}$ means the results of OPT and LLaMA-2 are not reported in the original OmniQuant paper. For LLaMA-1, LAMBADA and OpenbookQA are not included in OmniQuant either, thus we replace the results of these two tasks with FP16 results \textit{as an estimated upper limit} of OmniQuant. OmniQuant-r is the results we replicated using the official implementation\textsuperscript{\ref{footnote:s4:omniquant_code}} and checkpoints\textsuperscript{\ref{footnote:s4:omniquant_checkpoint}}. }
    \begin{center}
        \begin{small}
            \resizebox{\linewidth}{!}{
                \begin{tabular}{llcccccccccc}
                    \toprule
                    \multirow{2}{*}{Method} & \multirow{2}{*}{Q Config} & \multicolumn{3}{c}{OPT} & \multicolumn{3}{c}{LLaMA} & \multicolumn{3}{c}{LLaMA-2} & \multirow{2}{*}{ \makecell{Avg. $\Delta$ Accu.                                                                                              \\($\uparrow$)} }                                                                                                        \\
                    \cmidrule(lr){3-5} \cmidrule(lr){6-8} \cmidrule(lr){9-10}
                                            &                           & 6.7B                    & 13B                       & 30B                         & 7B                                             & 13B            & 33B            & 7B             & 13B            & 70B    &               \\ \midrule
                    FP16                    & -                         & 55.6\%                  & 56.2\%                    & 59.1\%                      & 63.2\%                                         & 65.0\%         & 68.4\%         & 63.5\%         & 66.5\%         & 69.9\% & -             \\ \midrule
                    GPTQ                    & INT4, g128                & \best{55.4\%}           & \good{56.4\%}             & -$^*$                       & 60.8\%                                         & 64.7\%         & 66.7\%         & 62.2\%         & \good{65.9\%}  & 69.8\% & -0.9\%        \\
                    AWQ                     & INT4, g128                & \good{55.3\%}           & \good{56.4\%}             & \best{58.9\%}               & 62.5\%                                         & \good{64.8\% } & \best{68.0\% } & 62.9\%         & \good{65.9\%}  & 69.9\% & \good{-0.4}\% \\
                    \texttt{LLM.int4()}     & $\tau=6.0$                & \best{55.4\%  }         & 55.9\%                    & 58.0\%                      & 62.2\%                                         & 64.6\%         & 67.7\%         & 62.6\%         & 65.8\%         & 69.9\% & -0.7\%        \\
                    OmniQuant$^\dag$        & W6A6, per-c/t             & -                       & -                         & -                           & 58.4\%                                         & 59.2\%         & 61.0\%         & -              & -              & -      & -6.0\%        \\
                    OmniQuant-r$^\dag$      & W6A6, per-c/t             & \best{55.4\%}           & 56.1\%                    & \good{58.6\%}               & 47.0\%                                         & 48.2\%         & 49.9\%         & 47.2\%         & 49.4\%         & 58.6   & -11.0\%       \\
                    \LRQa{}-\texttt{INT}    & W4A8, g128                & 54.1\%                  & 56.2\%                    & 57.7\%                      & 61.7\%                                         & 64.4\%         & 67.4\%         & 62.2\%         & \good{65.9\%}  & 69.7\% & -1.0\%        \\
                    \LRQa{}-\mxint{}        & W4A6                      & 54.7\%                  & 56.2\%                    & 58.5\%                      & \good{62.7\%}                                  & \best{64.9\%}  & \good{67.8\%}  & \good{63.0\%}  & 65.8\%         & 69.9\% & -0.5\%        \\
                    \LRQa{}-\mxint{}        & W4A8                      & 55.1\%                  & \best{56.5\%  }           & 58.4\%                      & \best{63.0\%  }                                & \good{64.8\% } & \best{68.0\% } & \best{63.1\% } & \best{66.1\% } & 69.9\% & \best{-0.3\%} \\ \bottomrule
                \end{tabular}
            }
        \end{small}
    \end{center}
    \label{tab:s4:w4a8-downstream}
\end{table*}


We present the perplexity ($\downarrow$), the average increased perplexity over models ($\Delta$ PPL ($\downarrow$)), average weight bitwidth, and circuit area ($\downarrow$) of \LRQa{} and existing $w$-only/$w\&a$ methods in~\Cref{tab:s4:w4a8-perplexity}. Then we exclude the methods with obvious performance degradation and evaluate the average downstream task performance in~\Cref{tab:s4:w4a8-downstream}. We additionally include a fixed-point version of ~\LRQa{} as a baseline. Best results in each setup are marked in \best{bold} and second best results are \good{underlined}.

\paragraph{WikiText-2} In the $w$-only quantization setup, \LRQa{}-\texttt{INT} achieves a significantly better perplexity when compared to GPTQ and is on par with AWQ (both only around $0.1$ higher than FP16), while has a substantially smaller hardware cost (circuit area).
In the $w\&a$ setup, \LRQa{}-\mxint{} has a perplexity around $0.15$ higher than FP16 when it is W4A8. \LRQa{}-\mxint{} outperforms state-of-the-art sub-8-bit methods by a significant margin. The perplexity of \LRQa{} is around $0.05$ higher than OmniQuant on the OPT family, but consistently outperforms OmniQuant on LLaMA family. Note that OmniQuant was trained on WikiText2 for 20 epochs~\cite{shao2023omniquant}, but \LRQa{} only proifles the activation magnitudes using 32 samples from a calibration dataset with Wikipedia texts excluded.

\paragraph{Downstream task accuracy}
We reuse the quantization setup in~\Cref{tab:s4:w4a8-perplexity} and conduct a thorough evaluation on downstream tasks, including ARC (easy), ARC (challenge), LAMBADA, PIQA, OpenBookQA and BoolQ and report the results in \Cref{tab:s4:w4a8-downstream}.
The average accuracy of \LRQa{} on the six downstream tasks is better than other quantization methods on LLaMA models, and nearly lossless (around 0.3\% drop) compared to the FP16 baseline. We reproduced the WikiText2 perplexity reported in OmniQuant paper~\cite{shao2023omniquant} using the official implementation\footnote{\label{footnote:s4:omniquant_code}\href{https://github.com/OpenGVLab/OmniQuant}{https://github.com/OpenGVLab/OmniQuant}} and checkpoints\footnote{\label{footnote:s4:omniquant_checkpoint}\href{https://huggingface.co/ChenMnZ/OmniQuant/tree/main}{https://huggingface.co/ChenMnZ/OmniQuant/tree/main}}, but failed to reproduce their downstream accuracy performance on LLaMA models. We refer to these mismatched OmniQuant results as OmniQuant-r in~\Cref{tab:s4:w4a8-downstream}.
We attribute the inconsistant behaviour of OmniQuant to its iterative quantization parameter training on WikiText2, which is further discussed in~\Cref{sec:appendix:omniquant}. Nevertheless, our method has demonstrated substantially better downstream task capabiliteis, with a much lower hardware cost (circuit area in \Cref{tab:s4:w4a8-perplexity}). A detailed discussion about hardware cost is in~\Cref{sec:appendix:hardware-cost}. A complete table including the accuracy of each individual task is in~\Cref{appendix:more-eval-results}.

\paragraph{AlpacaEval} We also evaluate the performance of \LRQa{} on AlpacaEval~\cite{alpaca_eval}, an evaluator for instruction-following language models. We use AlpacaEval to measure the fraction of times GPT-4 Turbo prefers the outputs from the quantized model over outputs from a reference model. Here we use AWQ as the reference model and report the results of LLaMA-2-7B-Chat and LLaMA-2-13B-Chat. We observe that \LRQa{} is competitive with AWQ in both length-controlled win rate and normal win rate.

\begin{table}[h]
    \centering
    \caption{AlpacaEval results. We use GPT-4 Turbo as the evaluator and AWQ as the reference model. The results are collected after evaluating LLaMA-2-7B-Chat/-13B-Chat on all samples. We find that \LRQa{} is competitive with AWQ in both length-controlled win rate and normal win rate.}
    \begin{small}
        \begin{tabular}{@{}cccc@{}}
            \toprule
            Model & Gen. vs Ref.                      & \makecell{Length-controlled            \\  win rate} & Win rate \\ \midrule
            7B    & \multirow{2}{*}{\LRQa{} vs AWQ} & 56.06 \%                    & 55.32 \% \\
            13B   &                                 & 52.90 \%                    & 52.51 \% \\ \bottomrule
        \end{tabular}
    \end{small}
    \label{tab:s4:alpaca-eval}
\end{table}

\paragraph{Hardware efficiency} \LRQa{} is more hardware friendly than the baselines. We highlight the last two columns of average weight bits and circuit area in~\Cref{tab:s4:w4a8-perplexity}. \LRQa{} requires less circuit area to implement a MACs when the model performance (perplexity and downstream task accuracy) and the MAC throughput are roughly matched with the baseline. We offer circuit area breakdowns of \texttt{LLM.int4()}, AWQ, and \LRQa{} in the~\Cref{tab:appendix:area-breakdown-llm-int4},~\Cref{tab:appendix:area-breakdown-awq}, and~\Cref{tab:appendix:area-breakdown-lrqa} in~\Cref{sec:appendix:hardware-cost}.

\paragraph{Optimization cost} The optimization of \LRQ{} is also efficient. The calibration and quantiation of LLaMA-33B takes around 1.2 hours in total on a single NVIDIA A100 GPU. In contrast, OmniQuant takes 7.3 hours to optimize the quantization parameters for LLaMA-33B. Furthermore, the optimization of \LRQ{} can be fully parallelized to be faster, since there is no dependency between the quantization of each linear layer such as fusing the scale matrices to preceding layers in SmoothQuant or knowledge distillation in LLM-QAT~\cite{liu2023llm}.

\paragraph{Other model families} To fully evaluate the adaptiveness of \LRQa{} across model families, we have also conducted experiments to evaluate its effectiveness on Vicuna and Mistral. Vicuna is an instruction-tuned LLaMA. Mistral uses Grouped-Query Attention (GQA)~\cite{ainslie2023gqa} and windowed attention~\cite{beltagy2020longformer}.
The results of Vicuna-v1.5-7B/13B and Mistral-7B are included in~\Cref{appendix:more-eval-results}. These results reveal a pattern consistent with other models, indicating that \LRQa{} is agnostic to various LLM families.


\subsection{$2$-bit Quantization}
\label{sec:experiments:2bit-quantization}

To explore the limit of \LRQa{}, we evaluate \LRQa{} in the 2-bit quantization setup. ~\Cref{tab:s4:w2a8-perplexity} compares \LRQa{} with OmniQuant and QuiP\#\footnote{\label{footnote:quip_sharp}QuiP\# is an improved version of QuiP released by the same research group: \href{https://github.com/Cornell-RelaxML/quip-sharp}{https://github.com/Cornell-RelaxML/quip-sharp}}, which are both recent works optimized for extremely low-precision LLM quantization.
We observe that 2-bit quantization is challenging for existing methods including \LRQa{}. These methods perform inconsistently with model sizes and families~(\Cref{tab:appendix:2-bit} in~\Cref{appendix:more-eval-results}). Unlike a simple rank $k=32$ for W4A8 quantization, \LRQa{} requires a larger $k$ for 2-bit quantization.

\begin{table}[ht]
  \caption{2-bit quantization perplexity ($\downarrow$) on WikiText2.
    OmniQuant and QuiP\#\textsuperscript{\ref{footnote:quip_sharp}} are two state-of-the-art methods for extremely low-precision LLM quantization.
    We found 2-bit quantization is still challenging for existing methods.}
  \begin{center}
    \begin{small}
      \begin{tabular}{lllcc}
        \toprule
        Q Setup                   & Method    & Q Config  & 7B           & 13B         \\ \midrule
        -                         & FP16      & -         & 5.67         & 5.10        \\ \midrule
        \multirow{3}{*}{$w$-only} & AWQ       & INT2 g128 & 2.6e5        & 2.8e5       \\
                                  & QuiP\#    & INT2 g128 & \good{10.97} & \good{8.43} \\
                                  & OmniQuant & INT2 g128 & 12.97        & 10.36       \\ \midrule
        $w\&a$                    & \LRQa{}   & $k=256$   & \best{10.30} & \best{8.42} \\ \bottomrule
      \end{tabular}
    \end{small}
  \end{center}
  \label{tab:s4:w2a8-perplexity}
\end{table}

\section{Conclusion}

In this work, we propose a novel LLM post-training quantization framework, \LRQ{}, which judiciously combine quantization and low-rank approximation to recover model capbility. We then further propose \LRQa{}, which leverages an activation-induced scale matrix to shape the singular values of quantization error towards a desirable distribution that can be accurate approximated. \LRQa{} achieves nearly-losses perplexity (around $0.15$ higher than FP16) and an average accuracy drop of only around $0.3\%$ on six different downstream tasks. The regular computation pattern of \LRQ{} ensures a higher hardware efficiency than existing methods and takes 67\% smaller circuit area than FP16.


\section*{Impact Statement}

This paper presents work whose goal is to advance the field of Machine Learning. There are many potential societal consequences of our work, none which we feel must be specifically highlighted here.



\bibliography{refs}

\begin{thebibliography}{53}
\providecommand{\natexlab}[1]{#1}
\providecommand{\url}[1]{\texttt{#1}}
\expandafter\ifx\csname urlstyle\endcsname\relax
  \providecommand{\doi}[1]{doi: #1}\else
  \providecommand{\doi}{doi: \begingroup \urlstyle{rm}\Url}\fi

\bibitem[Ainslie et~al.(2023)Ainslie, Lee-Thorp, de~Jong, Zemlyanskiy, Lebr{\'o}n, and Sanghai]{ainslie2023gqa}
Ainslie, J., Lee-Thorp, J., de~Jong, M., Zemlyanskiy, Y., Lebr{\'o}n, F., and Sanghai, S.
\newblock Gqa: Training generalized multi-query transformer models from multi-head checkpoints.
\newblock \emph{arXiv preprint arXiv:2305.13245}, 2023.

\bibitem[Beltagy et~al.(2020)Beltagy, Peters, and Cohan]{beltagy2020longformer}
Beltagy, I., Peters, M.~E., and Cohan, A.
\newblock Longformer: The long-document transformer.
\newblock \emph{arXiv preprint arXiv:2004.05150}, 2020.

\bibitem[Bisk et~al.(2020)Bisk, Zellers, Bras, Gao, and Choi]{Bisk2020}
Bisk, Y., Zellers, R., Bras, R.~L., Gao, J., and Choi, Y.
\newblock Piqa: Reasoning about physical commonsense in natural language.
\newblock In \emph{Thirty-Fourth AAAI Conference on Artificial Intelligence}, 2020.

\bibitem[Bondarenko et~al.(2021)Bondarenko, Nagel, and Blankevoort]{bondarenko2021understanding}
Bondarenko, Y., Nagel, M., and Blankevoort, T.
\newblock Understanding and overcoming the challenges of efficient transformer quantization.
\newblock \emph{arXiv preprint arXiv:2109.12948}, 2021.

\bibitem[Bondarenko et~al.(2023)Bondarenko, Nagel, and Blankevoort]{bondarenko2023quantizable}
Bondarenko, Y., Nagel, M., and Blankevoort, T.
\newblock Quantizable transformers: Removing outliers by helping attention heads do nothing.
\newblock \emph{arXiv preprint arXiv:2306.12929}, 2023.

\bibitem[Brown et~al.(2020)Brown, Mann, Ryder, Subbiah, Kaplan, Dhariwal, Neelakantan, Shyam, Sastry, Askell, et~al.]{brown2020language}
Brown, T., Mann, B., Ryder, N., Subbiah, M., Kaplan, J.~D., Dhariwal, P., Neelakantan, A., Shyam, P., Sastry, G., Askell, A., et~al.
\newblock Language models are few-shot learners.
\newblock \emph{Advances in neural information processing systems}, 33:\penalty0 1877--1901, 2020.

\bibitem[Chee et~al.(2023)Chee, Cai, Kuleshov, and De~Sa]{chee2023quip}
Chee, J., Cai, Y., Kuleshov, V., and De~Sa, C.
\newblock Quip: 2-bit quantization of large language models with guarantees.
\newblock \emph{arXiv preprint arXiv:2307.13304}, 2023.

\bibitem[Clark et~al.(2019)Clark, Lee, Chang, Kwiatkowski, Collins, and Toutanova]{clark2019boolq}
Clark, C., Lee, K., Chang, M.-W., Kwiatkowski, T., Collins, M., and Toutanova, K.
\newblock Boolq: Exploring the surprising difficulty of natural yes/no questions.
\newblock \emph{arXiv preprint arXiv:1905.10044}, 2019.

\bibitem[Clark et~al.(2018)Clark, Cowhey, Etzioni, Khot, Sabharwal, Schoenick, and Tafjord]{allenai_arc}
Clark, P., Cowhey, I., Etzioni, O., Khot, T., Sabharwal, A., Schoenick, C., and Tafjord, O.
\newblock Think you have solved question answering? try arc, the ai2 reasoning challenge.
\newblock \emph{arXiv:1803.05457v1}, 2018.

\bibitem[Darvish~Rouhani et~al.(2020)Darvish~Rouhani, Lo, Zhao, Liu, Fowers, Ovtcharov, Vinogradsky, Massengill, Yang, Bittner, et~al.]{darvish2020pushing}
Darvish~Rouhani, B., Lo, D., Zhao, R., Liu, M., Fowers, J., Ovtcharov, K., Vinogradsky, A., Massengill, S., Yang, L., Bittner, R., et~al.
\newblock Pushing the limits of narrow precision inferencing at cloud scale with microsoft floating point.
\newblock \emph{Advances in neural information processing systems}, 33:\penalty0 10271--10281, 2020.

\bibitem[Dettmers et~al.(2022)Dettmers, Lewis, Belkada, and Zettlemoyer]{dettmers2022llm}
Dettmers, T., Lewis, M., Belkada, Y., and Zettlemoyer, L.
\newblock Llm. int8 (): 8-bit matrix multiplication for transformers at scale.
\newblock \emph{arXiv preprint arXiv:2208.07339}, 2022.

\bibitem[Dettmers et~al.(2023{\natexlab{a}})Dettmers, Pagnoni, Holtzman, and Zettlemoyer]{dettmers2023qlora}
Dettmers, T., Pagnoni, A., Holtzman, A., and Zettlemoyer, L.
\newblock Qlora: Efficient finetuning of quantized llms.
\newblock \emph{arXiv preprint arXiv:2305.14314}, 2023{\natexlab{a}}.

\bibitem[Dettmers et~al.(2023{\natexlab{b}})Dettmers, Svirschevski, Egiazarian, Kuznedelev, Frantar, Ashkboos, Borzunov, Hoefler, and Alistarh]{dettmers2023spqr}
Dettmers, T., Svirschevski, R., Egiazarian, V., Kuznedelev, D., Frantar, E., Ashkboos, S., Borzunov, A., Hoefler, T., and Alistarh, D.
\newblock Spqr: A sparse-quantized representation for near-lossless llm weight compression.
\newblock \emph{arXiv preprint arXiv:2306.03078}, 2023{\natexlab{b}}.

\bibitem[Drumond et~al.(2018)Drumond, Lin, Jaggi, and Falsafi]{drumond2018training}
Drumond, M., Lin, T., Jaggi, M., and Falsafi, B.
\newblock Training dnns with hybrid block floating point.
\newblock \emph{Advances in Neural Information Processing Systems}, 31, 2018.

\bibitem[Fox et~al.(2020)Fox, Rasoulinezhad, Faraone, Leong, et~al.]{fox2020block}
Fox, S., Rasoulinezhad, S., Faraone, J., Leong, P., et~al.
\newblock A block minifloat representation for training deep neural networks.
\newblock In \emph{International Conference on Learning Representations}, 2020.

\bibitem[Frantar et~al.(2022)Frantar, Ashkboos, Hoefler, and Alistarh]{frantar2022gptq}
Frantar, E., Ashkboos, S., Hoefler, T., and Alistarh, D.
\newblock Gptq: Accurate post-training quantization for generative pre-trained transformers.
\newblock \emph{arXiv preprint arXiv:2210.17323}, 2022.

\bibitem[Gao et~al.(2023)Gao, Tow, Abbasi, Biderman, Black, DiPofi, Foster, Golding, Hsu, Le~Noac'h, Li, McDonell, Muennighoff, Ociepa, Phang, Reynolds, Schoelkopf, Skowron, Sutawika, Tang, Thite, Wang, Wang, and Zou]{eval-harness}
Gao, L., Tow, J., Abbasi, B., Biderman, S., Black, S., DiPofi, A., Foster, C., Golding, L., Hsu, J., Le~Noac'h, A., Li, H., McDonell, K., Muennighoff, N., Ociepa, C., Phang, J., Reynolds, L., Schoelkopf, H., Skowron, A., Sutawika, L., Tang, E., Thite, A., Wang, B., Wang, K., and Zou, A.
\newblock A framework for few-shot language model evaluation, 12 2023.
\newblock URL \url{https://zenodo.org/records/10256836}.

\bibitem[Hansen(2024)]{casper-hansen_autoawq}
Hansen, C.
\newblock Autoawq.
\newblock \url{https://github.com/casper-hansen/AutoAWQ}, 2024.

\bibitem[Hoffmann et~al.(2022)Hoffmann, Borgeaud, Mensch, Buchatskaya, Cai, Rutherford, Casas, Hendricks, Welbl, Clark, et~al.]{hoffmann2022training}
Hoffmann, J., Borgeaud, S., Mensch, A., Buchatskaya, E., Cai, T., Rutherford, E., Casas, D. d.~L., Hendricks, L.~A., Welbl, J., Clark, A., et~al.
\newblock Training compute-optimal large language models.
\newblock \emph{arXiv preprint arXiv:2203.15556}, 2022.

\bibitem[Hu et~al.(2021)Hu, Shen, Wallis, Allen-Zhu, Li, Wang, Wang, and Chen]{hu2021lora}
Hu, E.~J., Shen, Y., Wallis, P., Allen-Zhu, Z., Li, Y., Wang, S., Wang, L., and Chen, W.
\newblock Lora: Low-rank adaptation of large language models.
\newblock \emph{arXiv preprint arXiv:2106.09685}, 2021.

\bibitem[Jeon et~al.(2023)Jeon, Lee, Park, and Kim]{jeon2023frustratingly}
Jeon, Y., Lee, C., Park, K., and Kim, H.-y.
\newblock A frustratingly easy post-training quantization scheme for llms.
\newblock In \emph{Proceedings of the 2023 Conference on Empirical Methods in Natural Language Processing}, pp.\  14446--14461, 2023.

\bibitem[Jiang et~al.(2023)Jiang, Sablayrolles, Mensch, Bamford, Chaplot, Casas, Bressand, Lengyel, Lample, Saulnier, et~al.]{jiang2023mistral}
Jiang, A.~Q., Sablayrolles, A., Mensch, A., Bamford, C., Chaplot, D.~S., Casas, D. d.~l., Bressand, F., Lengyel, G., Lample, G., Saulnier, L., et~al.
\newblock Mistral 7b.
\newblock \emph{arXiv preprint arXiv:2310.06825}, 2023.

\bibitem[Lee et~al.(2023{\natexlab{a}})Lee, Kim, Baek, Hwang, Sung, and Choi]{lee2023enhancing}
Lee, J., Kim, M., Baek, S., Hwang, S.~J., Sung, W., and Choi, J.
\newblock Enhancing computation efficiency in large language models through weight and activation quantization.
\newblock \emph{arXiv preprint arXiv:2311.05161}, 2023{\natexlab{a}}.

\bibitem[Lee et~al.(2023{\natexlab{b}})Lee, Kim, Kwon, and Lee]{lee2022flexround}
Lee, J.~H., Kim, J., Kwon, S.~J., and Lee, D.
\newblock Flexround: Learnable rounding based on element-wise division for post-training quantization.
\newblock \emph{arXiv preprint arXiv:2306.00317}, 2023{\natexlab{b}}.

\bibitem[Li et~al.(2023{\natexlab{a}})Li, Zhang, Dubois, Taori, Gulrajani, Guestrin, Liang, and Hashimoto]{alpaca_eval}
Li, X., Zhang, T., Dubois, Y., Taori, R., Gulrajani, I., Guestrin, C., Liang, P., and Hashimoto, T.~B.
\newblock Alpacaeval: An automatic evaluator of instruction-following models.
\newblock \url{https://github.com/tatsu-lab/alpaca_eval}, 2023{\natexlab{a}}.

\bibitem[Li et~al.(2023{\natexlab{b}})Li, Yu, Liang, He, Karampatziakis, Chen, and Zhao]{li2023loftq}
Li, Y., Yu, Y., Liang, C., He, P., Karampatziakis, N., Chen, W., and Zhao, T.
\newblock Loftq: Lora-fine-tuning-aware quantization for large language models.
\newblock \emph{arXiv preprint arXiv:2310.08659}, 2023{\natexlab{b}}.

\bibitem[Lin et~al.(2023)Lin, Tang, Tang, Yang, Dang, and Han]{lin2023awq}
Lin, J., Tang, J., Tang, H., Yang, S., Dang, X., and Han, S.
\newblock Awq: Activation-aware weight quantization for llm compression and acceleration.
\newblock \emph{arXiv preprint arXiv:2306.00978}, 2023.

\bibitem[Lin(2024)]{noauthor_omniquant_nodate}
Lin, L.
\newblock {LLM-Tracker: OmniQuant}, 2024.
\newblock URL \url{https://llm-tracker.info/howto/OmniQuant}.

\bibitem[Liu et~al.(2023{\natexlab{a}})Liu, Gong, Wei, Dong, Cai, and Zhuang]{liu2023qllm}
Liu, J., Gong, R., Wei, X., Dong, Z., Cai, J., and Zhuang, B.
\newblock Qllm: Accurate and efficient low-bitwidth quantization for large language models.
\newblock \emph{arXiv preprint arXiv:2310.08041}, 2023{\natexlab{a}}.

\bibitem[Liu et~al.(2023{\natexlab{b}})Liu, Oguz, Zhao, Chang, Stock, Mehdad, Shi, Krishnamoorthi, and Chandra]{liu2023llm}
Liu, Z., Oguz, B., Zhao, C., Chang, E., Stock, P., Mehdad, Y., Shi, Y., Krishnamoorthi, R., and Chandra, V.
\newblock Llm-qat: Data-free quantization aware training for large language models.
\newblock \emph{arXiv preprint arXiv:2305.17888}, 2023{\natexlab{b}}.

\bibitem[Luccioni et~al.(2023)Luccioni, Viguier, and Ligozat]{luccioni2023estimating}
Luccioni, A.~S., Viguier, S., and Ligozat, A.-L.
\newblock Estimating the carbon footprint of bloom, a 176b parameter language model.
\newblock \emph{Journal of Machine Learning Research}, 24\penalty0 (253):\penalty0 1--15, 2023.

\bibitem[Luo et~al.(2023)Luo, Gao, Zhang, Fan, Zhang, and Xu]{luo2023long}
Luo, Y., Gao, Y., Zhang, Z., Fan, J., Zhang, H., and Xu, M.
\newblock Long-range zero-shot generative deep network quantization.
\newblock \emph{Neural Networks}, 166:\penalty0 683--691, 2023.

\bibitem[Marchenko \& Pastur(1967)Marchenko and Pastur]{marchenko1967}
Marchenko, V. and Pastur, L.
\newblock Distribution of eigenvalues for some sets of random matrices.
\newblock \emph{Mat. Sb}, 72:\penalty0 507--536, 1967.

\bibitem[Merity et~al.(2016)Merity, Xiong, Bradbury, and Socher]{merity2016pointer}
Merity, S., Xiong, C., Bradbury, J., and Socher, R.
\newblock Pointer sentinel mixture models, 2016.

\bibitem[Micikevicius et~al.(2023)Micikevicius, Oberman, Dubey, Cornea, Rodriguez, Bratt, Grisenthwaite, Jouppi, Chou, Huffman, Schulte, Wittig, Jani, and Deng]{mxspecs2023}
Micikevicius, P., Oberman, S., Dubey, P., Cornea, M., Rodriguez, A., Bratt, I., Grisenthwaite, R., Jouppi, N., Chou, C., Huffman, A., Schulte, M., Wittig, R., Jani, D., and Deng, S.
\newblock Ocp 8-bit floating point specification (ofp8), 2023.
\newblock URL \url{https://www.opencompute.org/documents/ocp-microscaling-formats-mx-v1-0-spec-final-pdf}.

\bibitem[Mihaylov et~al.(2018)Mihaylov, Clark, Khot, and Sabharwal]{OpenBookQA2018}
Mihaylov, T., Clark, P., Khot, T., and Sabharwal, A.
\newblock Can a suit of armor conduct electricity? a new dataset for open book question answering.
\newblock In \emph{EMNLP}, 2018.

\bibitem[Nagel et~al.(2021)Nagel, Fournarakis, Amjad, Bondarenko, Van~Baalen, and Blankevoort]{nagel2021white}
Nagel, M., Fournarakis, M., Amjad, R.~A., Bondarenko, Y., Van~Baalen, M., and Blankevoort, T.
\newblock A white paper on neural network quantization.
\newblock \emph{arXiv preprint arXiv:2106.08295}, 2021.

\bibitem[Paperno et~al.(2016)Paperno, Kruszewski, Lazaridou, Pham, Bernardi, Pezzelle, Baroni, Boleda, and Fern{\'a}ndez]{paperno2016lambada}
Paperno, D., Kruszewski, G., Lazaridou, A., Pham, Q.~N., Bernardi, R., Pezzelle, S., Baroni, M., Boleda, G., and Fern{\'a}ndez, R.
\newblock The lambada dataset: Word prediction requiring a broad discourse context.
\newblock \emph{arXiv preprint arXiv:1606.06031}, 2016.

\bibitem[Rouhani et~al.(2023{\natexlab{a}})Rouhani, Zhao, Elango, Shafipour, Hall, Mesmakhosroshahi, More, Melnick, Golub, Varatkar, et~al.]{rouhani2023shared}
Rouhani, B., Zhao, R., Elango, V., Shafipour, R., Hall, M., Mesmakhosroshahi, M., More, A., Melnick, L., Golub, M., Varatkar, G., et~al.
\newblock Shared microexponents: A little shifting goes a long way.
\newblock \emph{arXiv preprint arXiv:2302.08007}, 2023{\natexlab{a}}.

\bibitem[Rouhani et~al.(2023{\natexlab{b}})Rouhani, Zhao, More, Hall, Khodamoradi, Deng, Choudhary, Cornea, Dellinger, Denolf, et~al.]{rouhani2023microscaling}
Rouhani, B.~D., Zhao, R., More, A., Hall, M., Khodamoradi, A., Deng, S., Choudhary, D., Cornea, M., Dellinger, E., Denolf, K., et~al.
\newblock Microscaling data formats for deep learning.
\newblock \emph{arXiv preprint arXiv:2310.10537}, 2023{\natexlab{b}}.

\bibitem[Shao et~al.(2023)Shao, Chen, Zhang, Xu, Zhao, Li, Zhang, Gao, Qiao, and Luo]{shao2023omniquant}
Shao, W., Chen, M., Zhang, Z., Xu, P., Zhao, L., Li, Z., Zhang, K., Gao, P., Qiao, Y., and Luo, P.
\newblock Omniquant: Omnidirectionally calibrated quantization for large language models.
\newblock \emph{arXiv preprint arXiv:2308.13137}, 2023.

\bibitem[Soboleva et~al.(2023)Soboleva, Al-Khateeb, Myers, Steeves, Hestness, and Dey]{cerebras2023slimpajama}
Soboleva, D., Al-Khateeb, F., Myers, R., Steeves, J.~R., Hestness, J., and Dey, N.
\newblock {SlimPajama: A 627B token cleaned and deduplicated version of RedPajama}.
\newblock \url{https://www.cerebras.net/blog/slimpajama-a-627b-token-cleaned-and-deduplicated-version-of-redpajama}, 2023.
\newblock URL \url{https://huggingface.co/datasets/cerebras/SlimPajama-627B}.

\bibitem[Tang et~al.(2023)Tang, Sun, Wu, Liu, Zhu, and Kang]{tang2023easyquant}
Tang, H., Sun, Y., Wu, D., Liu, K., Zhu, J., and Kang, Z.
\newblock Easyquant: An efficient data-free quantization algorithm for llms.
\newblock In \emph{Proceedings of the 2023 Conference on Empirical Methods in Natural Language Processing}, pp.\  9119--9128, 2023.

\bibitem[Touvron et~al.(2023{\natexlab{a}})Touvron, Lavril, Izacard, Martinet, Lachaux, Lacroix, Rozi{\`e}re, Goyal, Hambro, Azhar, et~al.]{touvron2023llama}
Touvron, H., Lavril, T., Izacard, G., Martinet, X., Lachaux, M.-A., Lacroix, T., Rozi{\`e}re, B., Goyal, N., Hambro, E., Azhar, F., et~al.
\newblock Llama: Open and efficient foundation language models.
\newblock \emph{arXiv preprint arXiv:2302.13971}, 2023{\natexlab{a}}.

\bibitem[Touvron et~al.(2023{\natexlab{b}})Touvron, Martin, Stone, Albert, Almahairi, Babaei, Bashlykov, Batra, Bhargava, Bhosale, et~al.]{touvron2023llama2}
Touvron, H., Martin, L., Stone, K., Albert, P., Almahairi, A., Babaei, Y., Bashlykov, N., Batra, S., Bhargava, P., Bhosale, S., et~al.
\newblock Llama 2: Open foundation and fine-tuned chat models.
\newblock \emph{arXiv preprint arXiv:2307.09288}, 2023{\natexlab{b}}.

\bibitem[Wei et~al.(2022)Wei, Zhang, Zhang, Gong, Zhang, Zhang, Yu, and Liu]{wei2022outlier}
Wei, X., Zhang, Y., Zhang, X., Gong, R., Zhang, S., Zhang, Q., Yu, F., and Liu, X.
\newblock Outlier suppression: Pushing the limit of low-bit transformer language models.
\newblock \emph{Advances in Neural Information Processing Systems}, 35:\penalty0 17402--17414, 2022.

\bibitem[Wei et~al.(2023)Wei, Zhang, Li, Zhang, Gong, Guo, and Liu]{wei2023outlier}
Wei, X., Zhang, Y., Li, Y., Zhang, X., Gong, R., Guo, J., and Liu, X.
\newblock Outlier suppression+: Accurate quantization of large language models by equivalent and optimal shifting and scaling.
\newblock \emph{arXiv preprint arXiv:2304.09145}, 2023.

\bibitem[Workshop et~al.(2022)Workshop, Scao, Fan, Akiki, Pavlick, Ili{\'c}, Hesslow, Castagn{\'e}, Luccioni, Yvon, et~al.]{workshop2022bloom}
Workshop, B., Scao, T.~L., Fan, A., Akiki, C., Pavlick, E., Ili{\'c}, S., Hesslow, D., Castagn{\'e}, R., Luccioni, A.~S., Yvon, F., et~al.
\newblock Bloom: A 176b-parameter open-access multilingual language model.
\newblock \emph{arXiv preprint arXiv:2211.05100}, 2022.

\bibitem[Xiao et~al.(2023)Xiao, Lin, Seznec, Wu, Demouth, and Han]{xiao2023smoothquant}
Xiao, G., Lin, J., Seznec, M., Wu, H., Demouth, J., and Han, S.
\newblock Smoothquant: Accurate and efficient post-training quantization for large language models.
\newblock In \emph{International Conference on Machine Learning}, pp.\  38087--38099. PMLR, 2023.

\bibitem[Zhang et~al.(2023)Zhang, Cheng, Shumailov, Constantinides, and Zhao]{zhang-etal-2023-revisiting}
Zhang, C., Cheng, J., Shumailov, I., Constantinides, G., and Zhao, Y.
\newblock Revisiting block-based quantisation: What is important for sub-8-bit {LLM} inference?
\newblock In Bouamor, H., Pino, J., and Bali, K. (eds.), \emph{Proceedings of the 2023 Conference on Empirical Methods in Natural Language Processing}, pp.\  9988--10006, Singapore, December 2023. Association for Computational Linguistics.
\newblock \doi{10.18653/v1/2023.emnlp-main.617}.
\newblock URL \url{https://aclanthology.org/2023.emnlp-main.617}.

\bibitem[Zhang et~al.(2022{\natexlab{a}})Zhang, Roller, Goyal, Artetxe, Chen, Chen, Dewan, Diab, Li, Lin, et~al.]{zhang2022opt}
Zhang, S., Roller, S., Goyal, N., Artetxe, M., Chen, M., Chen, S., Dewan, C., Diab, M., Li, X., Lin, X.~V., et~al.
\newblock Opt: Open pre-trained transformer language models.
\newblock \emph{arXiv preprint arXiv:2205.01068}, 2022{\natexlab{a}}.

\bibitem[Zhang et~al.(2022{\natexlab{b}})Zhang, McDanel, and Kung]{zhang2022fast}
Zhang, S.~Q., McDanel, B., and Kung, H.
\newblock Fast: Dnn training under variable precision block floating point with stochastic rounding.
\newblock In \emph{2022 IEEE International Symposium on High-Performance Computer Architecture (HPCA)}, pp.\  846--860. IEEE, 2022{\natexlab{b}}.

\bibitem[Zheng et~al.(2023)Zheng, Chiang, Sheng, Zhuang, Wu, Zhuang, Lin, Li, Li, Xing, et~al.]{zheng2023judging}
Zheng, L., Chiang, W.-L., Sheng, Y., Zhuang, S., Wu, Z., Zhuang, Y., Lin, Z., Li, Z., Li, D., Xing, E., et~al.
\newblock Judging llm-as-a-judge with mt-bench and chatbot arena.
\newblock \emph{arXiv preprint arXiv:2306.05685}, 2023.

\end{thebibliography}
\bibliographystyle{icml2024}

\newpage
\appendix

\section{Data Calibration}
\label{sec:appendix:data-calibration}


Given a calibration dataset containing $N$ samples, $\{X_i \mid i=1,2,\dots,N\}$, we first profile the activation magnitude for each channel,
\begin{equation}
    \begin{split}
        \mathbf{a}_i     & = \text{mean}(|X_i|, \text{axis}=0), \\
        \mathbf{\bar{a}} & = \max(
        \begin{bmatrix}
            \mathbf{a}_1 \\
            \vdots       \\
            \mathbf{a}_N
        \end{bmatrix}, \text{axis}=0),
    \end{split}
    \label{eq:appendix:calibration_1}
\end{equation}
where $|\cdot|$ calculates the element-wise absolute value and $\mathbf{\bar{a}}=[a_1, a_2, \dots, a_m]$ is a row vector of maximum channel magnitudes across samples. We normalize $\mathbf{\bar{a}}$ to get the diagonal matrix $S$:
\begin{equation}
    s_i=\frac{a_i}{\sqrt{\min(\mathbf{\bar{a}}) \times \max(\mathbf{\bar{a}})}},
    \label{eq:appendix:calibration_2}
\end{equation}
\Cref{eq:appendix:calibration_1} and~\Cref{eq:appendix:calibration_2} are empirical implementation based on~\cite{lin2023awq}. We leave the exploration of an analytical derivation of $S$ as future work.


\section{Comparison between \LRQn{} and \LRQa{}}
\label{sec:appendix:svd-distribution-comparison}

\begin{figure}[ht]
    \begin{center}
        \includegraphics[width=0.5\textwidth]{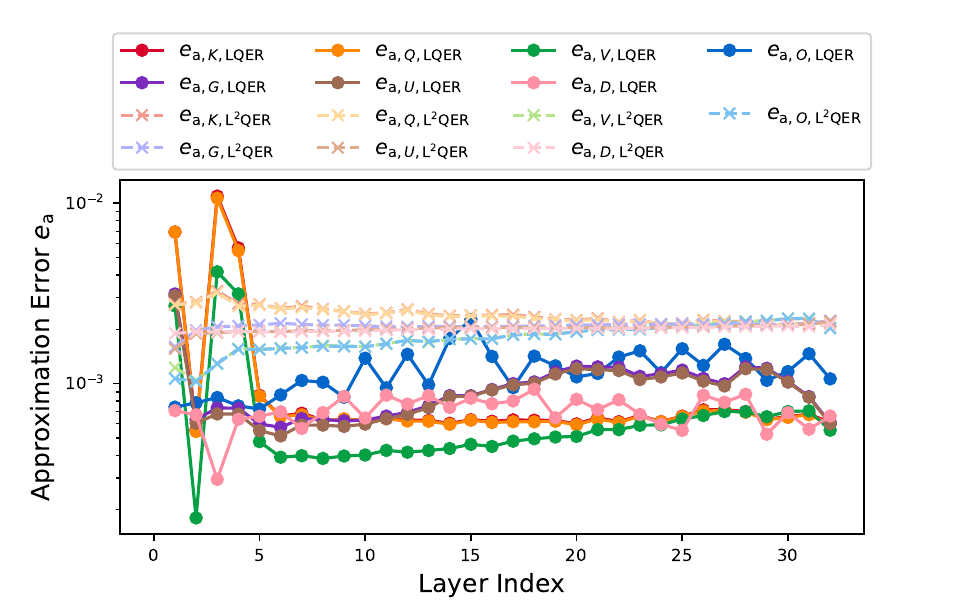}
        \vskip 0.1in
        \caption{Approximation error of \LRQn{} and \LRQa{} across decoder layers in LLaMA-7B. \LRQa{} produces smaller approximation errors on most of the linear layers in transformer-based LLMs. However, there are a few layers better reconstructed by \LRQn{}, such as the key, value, output project layers in 1st, 3rd, and 4th decoder layer. The derivation of $S$ worths further exploration.}
        \label{fig:appendix:q-approximation-error-vs-layer-id}
    \end{center}
    \vskip 0.1in
\end{figure}

Here we visualzie the approximation error of \LRQ{} and \LRQn{} versus layer index in~\Cref{fig:appendix:q-approximation-error-vs-layer-id}. The approximation error is measured as:
\begin{equation}
    e_a=\frac{1}{mn}\sum_{i=1}^{m}\sum_{j=1}^{n}(|E_q-\widetilde{E}_q|_{i,j})
\end{equation}
where $|\cdot|$ calculate the element-wise absolute value. \LRQa{} reconstructs the quantization error more accurate than \LRQn{} on most layers, while \LRQn{} better reconstruct the K, Q, and V projection layers at the 1st, 3rd, and 4th transformer layers.

\section{Inconsistant performance of OmniQuant on WikiText2 and downstream tasks}
\label{sec:appendix:omniquant}

OmniQuant is one of the state-of-the-art LLM post-training-quantization methods we compared in this work. Thanks for the official open-sourced implementation and quantization parameter checkpoints, we performed extensive experiments to compare OmniQuant to \LRQ{}. We sucessfully reproduce the perplexity and downstream task accuracy of OPT-family. However, the LLaMA models quantized by OmniQuant have obvious performance degradation on downstream tasks, around 18.9\% lower than FP16 baselines on average.

We attribute this performance degradation to the iterative gradient-base training on WikiText2 in OmniQuant. As stated in~\cite{shao2023omniquant}, OmniQuant optimizes the quantization parameter (shifts and scales) by training on WikiText2 samples for 20 epochs (40 epochs for W2A16). This training requires tuning the hyper-parameters such as number of training samples, learning rates and total number of epochs, which may cause overfitting or underfitting if not tuned properly. Both cases can be the reason for performance degradation.

\section{Estimate Hardware Cost}
\label{sec:appendix:hardware-cost}


We estimate the memory efficiency with average bitwidth. The average bitwidth of per-channel scaled quantization is considered as the average bits of an FP16 scalor and $m$ fixed-point numbers, where $m$ is the input hidden size. The average bitwidth of \mxint{} is averaged across one shared exponent and $B$ mantissas, where $B$ is the block size. For \LRQ{}/\LRQa{}, this is averaged across the low-precision $W_q$ and the high-precision $A_k$ and $B_k$. The average weight bitwidth of \LRQa{} in memory is 0.2 higher than GPTQ and AWQ, which is mainly contributed by the two low-rank matrices $A_k$ and $B_k$. \LRQa{} outperforms existing nearly lossless methods in terms of circuit area, because it is free from expensive element-wise dequantization (GPTQ and AWQ), or scatter/gather operations (\texttt{LLM.int4()}) at runtime. 

We estimate the hardware cost with circuit area. We mapped the algorithms of these approaches onto custom hardware accelerators on FPGAs. To ensure fairness, these hardware accelerators have the same throughput of 16 multiply-accumulate (MAC) operations per clock cycle when computing a linear operation of the same matrix sizes. We then measure the circuit area in terms of LUTs and Digital Signal Processing blocks (DSPs) on the FPGA, where a DSP is treated as 100 LUTs. The area results were measured from the Place \& Route report in Xilinx Vivado 2023.1. The FPGA family that we used for all the experiments is Xilinx Alveo U250. We summarize the area breakdown of \texttt{LLM.int4()}, AWQ, and \LRQa{} in~\Cref{tab:appendix:area-breakdown-llm-int4},~\Cref{tab:appendix:area-breakdown-awq}, and~\Cref{tab:appendix:area-breakdown-lrqa}, respectively.

\begin{table}[h]
    \centering
    \caption{Area breakdown of \texttt{LLM.int4()}, where $\text{GEMM}_l$ and $\text{GEMM}_h$ are low-precision and high-precision GEMM operations respectively.}
    \begin{small}
        \begin{tabular}{@{}lcccc@{}}
            \toprule
            \texttt{LLM.int4()} & \makecell{$\text{GEMM}_l$+                          \\ casting} & \makecell{Scatter +                  \\ gather} & $\text{GEMM}_h$ & Other  \\ \midrule
            LUTs                & 106959                     & 11579 & 404   & 13604  \\
            Percentage          & 80.7 \%                    & 8.8\% & 3.0\% & 10.3\% \\ \bottomrule
        \end{tabular}
    \end{small}
    \label{tab:appendix:area-breakdown-llm-int4}
\end{table}

\begin{table}[h]
    \centering
    \caption{Area breakdown of AWQ}
    \begin{small}
        \begin{tabular}{@{}lccc@{}}
            \toprule
            AWQ        & Dequantize & Matmul & Other  \\ \midrule
            LUTs       & 62907      & 11476  & 11131  \\
            Percentage & 73.6\%     & 13.4\% & 13.0\% \\ \bottomrule
        \end{tabular}
    \end{small}
    \label{tab:appendix:area-breakdown-awq}
\end{table}

\begin{table}[h]
    \centering
    \caption{Area breakdown of \LRQa{}, where Matmul1, Matmul2, and Matmul3 are $XW_q$, $X A_k$, and $(XA_k)B_k$ respectively.}
    \begin{small}
        \begin{tabular}{@{}lccc@{}}
            \toprule
            \LRQa{}    & Matmul2 & Matmul1 & Matmul3 \\ \midrule
            LUTs       & 1782    & 1028    & 992     \\
            Percentage & 34.5\%  & 19.9\%  & 19.2\%  \\ \bottomrule
        \end{tabular}
    \end{small}
    \label{tab:appendix:area-breakdown-lrqa}
\end{table}

\section{More evaluation results}
\label{appendix:more-eval-results}

We present the complete results of each specific downstream tasks in~\Cref{tab:appendix:opt-6.7b,tab:appendix:opt-13b,tab:appendix:opt-30b,tab:appendix:llama-7b,tab:appendix:llama-13b,tab:appendix:LLaMA-30b,tab:appendix:llama-2-7b,tab:appendix:llama-2-13b}. We also tested \LRQa{} on Vicuna-7b/13b and Mistral-7b-v0.1 in~\Cref{tab:appendix:vicuna-7b,tab:appendix:Vicuna-13b,tab:appendix:mistral-7b}.

\begin{table}[t]
\centering
\caption{More 2-bit $w$-only results. These methods perform inconsistently with model sizes and families. Unlike a simple rank $k=32$ for W4A8 quantization, \LRQa{}requires a larger rank $k$ for 2-bit quantization.}
\begin{small}
\begin{tabular}{@{}lccccc@{}}
\toprule
\multirow{2}{*}{Method} & \multicolumn{3}{c}{OPT}                          & \multicolumn{2}{c}{LLaMA}      \\ \cmidrule(l){2-6} 
                        & 125M           & 1.3B           & 2.7B           & 7B             & 13B           \\ \midrule 
FP16                    & 27.65          & 14.63          & 12.47          & 5.67           & 5.10          \\ \midrule
OmniQuant               & \good{75.43}    & \best{23.95} & \best{18.13} & 12.97          & 10.36         \\
Quip                    & 347.40         & 41.64          & 2998.00        & \good{ 10.97}    & \good{8.43}    \\
\LRQa{}                   & \best{45.29} & \good{29.82}    & \good{23.76}    & \best{10.30} & \best{8.42} \\ \bottomrule
\end{tabular}
\end{small}
\label{tab:appendix:2-bit}
\end{table}
\newpage

\begin{table*}[t]
\centering
\caption{OPT-6.7B}
\vskip 0.1in
\begin{small}
\resizebox{\linewidth}{!}{
\begin{tabular}{@{}lcccccccc@{}}
\toprule
Method & WikiText2 & ARC (easy) & ARC (challenge) & LAMBADA & PIQA & BOOLQ & OpenbookQA & Avg. Accuracy \\ \midrule
FP16 & 10.86 & 65.6\% & 30.5\% & 67.7\% & 76.3\% & 66.1\% & 27.6\% & 55.6\% \\
GPTQ & 10.95 & 65.6\% & 31.1\% & 68.5\% & 76.2\% & 65.2\% & 26.2\% & 55.4\% \\
AWQ & 10.93 & 65.3\% & 30.5\% & 67.4\% & 76.6\% & 65.2\% & 26.6\% & 55.3\% \\
LLM.int4() & 11.23 & 65.3\% & 30.5\% & 67.4\% & 76.6\% & 65.2\% & 26.6\% & 55.3\% \\
OmniQuant (W6A6) & 10.96 & 65.4\% & 30.9\% & 66.9\% & 76.0\% & 66.2\% & 26.8\% & 55.4\% \\
LQER-INT (W4A8) & 11.10 & 63.8\% & 29.6\% & 65.7\% & 75.6\% & 63.1\% & 26.8\% & 54.1\% \\
LQER-MXINT (W4A6) & 11.03 & 65.4\% & 30.5\% & 65.6\% & 75.4\% & 64.0\% & 27.6\% & 54.7\% \\
LQER-MXINT (W4A8) & 11.00 & 65.2\% & 30.4\% & 66.3\% & 75.5\% & 65.3\% & 27.6\% & 55.0\% \\ \bottomrule
\end{tabular}
}
\end{small}
\vskip 0.1in
\label{tab:appendix:opt-6.7b}
\end{table*}

\begin{table*}[t]
\centering
\caption{OPT-13B}
\vskip 0.1in
\begin{small}
\resizebox{\linewidth}{!}{
\begin{tabular}{@{}lcccccccc@{}}
\toprule
Method & WikiText2 & ARC (easy) & ARC (challenge) & LAMBADA & PIQA & BOOLQ & OpenbookQA & Avg. Accuracy \\ \midrule
FP16 & 10.13 & 67.1\% & 32.9\% & 68.6\% & 76.0\% & 65.8\% & 27.0\% & 56.2\% \\
GPTQ & 10.31 & 67.5\% & 32.8\% & 68.8\% & 76.1\% & 65.9\% & 27.2\% & 56.4\% \\
AWQ & 10.21 & 66.8\% & 33.3\% & 68.2\% & 75.6\% & 66.5\% & 28.0\% & 56.4\% \\
LLM.int4() & 10.39 & 66.2\% & 33.6\% & 67.8\% & 76.2\% & 67.3\% & 24.2\% & 55.9\% \\
OmniQuant (W6A6) & 10.96 & 67.1\% & 33.1\% & 68.4\% & 76.2\% & 65.3\% & 26.4\% & 56.1\% \\
LQER-INT (W4A8) & 10.38 & 66.5\% & 33.2\% & 67.5\% & 75.5\% & 67.9\% & 26.4\% & 56.2\% \\
LQER-MXINT (W4A6) & 10.32 & 67.2\% & 32.2\% & 67.9\% & 75.7\% & 68.3\% & 25.8\% & 56.2\% \\
LQER-MXINT (W4A8) & 10.27 & 67.4\% & 32.6\% & 68.4\% & 76.1\% & 68.3\% & 26.2\% & 56.5\% \\ \bottomrule
\end{tabular}
}
\end{small}
\vskip 0.1in
\label{tab:appendix:opt-13b}
\end{table*}

\begin{table*}[t]
\centering
\caption{OPT-6.7B}
\vskip 0.1in
\begin{small}
\resizebox{\linewidth}{!}{
\begin{tabular}{@{}lcccccccc@{}}
\toprule
Method & WikiText2 & ARC (easy) & ARC (challenge) & LAMBADA & PIQA & BOOLQ & OpenbookQA & Avg. Accuracy \\ \midrule
FP16 & 9.56 & 70.0\% & 34.6\% & 71.5\% & 77.6\% & 70.5\% & 30.2\% & 59.1\% \\
GPTQ & 9.63 & 62.2\% & 29.4\% & 74.9\% & 67.6\% & 69.1\% & 23.8\% & 54.5\% \\
AWQ & 9.59 & 69.7\% & 34.6\% & 71.6\% & 77.3\% & 70.4\% & 30.0\% & 58.9\% \\
LLM.int4() & 10.01 & 69.0\% & 32.8\% & 71.3\% & 76.9\% & 70.2\% & 27.8\% & 58.0\% \\
OmniQuant (W6A6) & 9.62 & 70.1\% & 34.2\% & 70.4\% & 77.3\% & 70.2\% & 29.6\% & 58.6\% \\
LQER-INT (W4A8) & 9.72 & \multicolumn{1}{l}{} & \multicolumn{1}{l}{} & \multicolumn{1}{l}{} & \multicolumn{1}{l}{} & \multicolumn{1}{l}{} & \multicolumn{1}{l}{} & \multicolumn{1}{l}{} \\
LQER-MXINT (W4A6) & 9.72 & 0.6990740741 & 0.3421501706 & 0.7050261983 & 0.7725788901 & 0.6923547401 & 0.298 & 58.5\% \\
LQER-MXINT (W4A8) & 9.67 & 69.4\% & 34.4\% & 70.4\% & 77.3\% & 69.5\% & 29.6\% & 58.4\% \\ \bottomrule
\end{tabular}
}
\end{small}
\vskip 0.1in
\label{tab:appendix:opt-30b}
\end{table*}

\begin{table*}[t]
\centering
\caption{llama-7B}
\vskip 0.1in
\begin{small}
\resizebox{\linewidth}{!}{
\begin{tabular}{@{}lcccccccc@{}}
\toprule
Method & WikiText2 & ARC (easy) & ARC (challenge) & LAMBADA & PIQA & BOOLQ & OpenbookQA & Avg. Accuracy \\ \midrule
FP16 & 5.10 & 77.4\% & 46.4\% & 76.2\% & 79.1\% & 78.0\% & 33.2\% & 65.0\% \\
GPTQ & 5.21 & 76.9\% & 46.8\% & 75.0\% & 79.3\% & 76.4\% & 34.0\% & 64.7\% \\
AWQ & 5.20 & 77.2\% & 46.4\% & 75.6\% & 79.0\% & 77.8\% & 32.8\% & 64.8\% \\
LLM.int4() & 5.31 & 77.2\% & 46.0\% & 75.4\% & 78.9\% & 77.1\% & 32.8\% & 64.6\% \\
OmniQuant (W6A6) & 5.28 & 72.5\% & 42.9\% & 0.0\% & 78.2\% & 66.4\% & 29.0\% & 48.2\% \\
LQER-INT (W4A8) & 5.31 & 76.9\% & 45.9\% & 74.0\% & 78.7\% & 77.2\% & 33.6\% & 64.4\% \\
LQER-MXINT (W4A6) & 5.24 & 77.1\% & 46.2\% & 75.6\% & 79.2\% & 77.6\% & 33.6\% & 64.9\% \\
LQER-MXINT (W4A8) & 5.21 & 77.0\% & 46.3\% & 75.6\% & 79.6\% & 77.3\% & 33.2\% & 64.8\% \\ \bottomrule
\end{tabular}
}
\end{small}
\vskip 0.1in
\label{tab:appendix:llama-7b}
\end{table*}

\begin{table*}[t]
\centering
\caption{LLaMA-13B}
\vskip 0.1in
\begin{small}
\resizebox{\linewidth}{!}{
\begin{tabular}{@{}lcccccccc@{}}
\toprule
Method & WikiText2 & ARC (easy) & ARC (challenge) & LAMBADA & PIQA & BOOLQ & OpenbookQA & Avg. Accuracy \\ \midrule
FP16 & 5.67 & 75.4\% & 41.9\% & 73.5\% & 78.7\% & 75.1\% & 34.4\% & 63.2\% \\
GPTQ & 9.63 & 73.6\% & 40.4\% & 70.0\% & 77.7\% & 73.0\% & 30.0\% & 60.8\% \\
AWQ & 9.59 & 75.5\% & 41.1\% & 72.5\% & 78.6\% & 74.9\% & 32.2\% & 62.5\% \\
LLM.int4() & 10.01 & 74.6\% & 42.1\% & 70.3\% & 78.6\% & 74.8\% & 32.8\% & 62.2\% \\
OmniQuant (W6A6) & 9.62 & 66.4\% & 38.8\% & 0.0\% & 76.7\% & 72.8\% & 27.2\% & 47.0\% \\
LQER-INT (W4A8) & 6.09 & 73.9\% & 40.6\% & 73.4\% & 77.7\% & 74.0\% & 30.6\% & 61.7\% \\
LQER-MXINT (W4A6) & 5.92 & 74.8\% & 41.5\% & 73.4\% & 78.2\% & 75.2\% & 33.0\% & 62.7\% \\
LQER-MXINT (W4A8) & 5.89 & 74.9\% & 41.6\% & 73.3\% & 78.6\% & 76.1\% & 33.6\% & 63.0\% \\ \bottomrule
\end{tabular}
}
\end{small}
\vskip 0.1in
\label{tab:appendix:llama-13b}
\end{table*}

\begin{table*}[t]
\centering
\caption{LLaMA-30B}
\vskip 0.1in
\begin{small}
\resizebox{\linewidth}{!}{
\begin{tabular}{@{}lcccccccc@{}}
\toprule
Method & WikiText2 & ARC (easy) & ARC (challenge) & LAMBADA & PIQA & BOOLQ & OpenbookQA & Avg. Accuracy \\ \midrule
FP16 & 4.10 & 80.4\% & 52.8\% & 77.6\% & 81.1\% & 82.7\% & 36.0\% & 68.4\% \\
GPTQ & 4.24 & 80.7\% & 50.2\% & 77.6\% & 80.5\% & 83.1\% & 35.8\% & 68.0\% \\
AWQ & 4.22 & 74.1\% & 46.0\% & 0.0\% & 79.5\% & 68.3\% & 31.4\% & 49.9\% \\
LLM.int4() & 4.33 & 79.0\% & 48.9\% & 75.8\% & 80.2\% & 82.4\% & 33.6\% & 66.7\% \\
OmniQuant (W6A6) & 4.38 & 74.1\% & 46.0\% & 0.0\% & 79.5\% & 68.3\% & 31.4\% & 49.9\% \\
LQER-INT (W4A8) & 4.35 & 80.1\% & 49.7\% & 77.0\% & 80.7\% & 81.5\% & 35.2\% & 67.4\% \\
LQER-MXINT (W4A6) & 4.28 & 80.1\% & 50.9\% & 77.4\% & 80.6\% & 82.4\% & 35.4\% & 67.8\% \\
LQER-MXINT (W4A8) & 4.25 & 80.0\% & 50.8\% & 77.6\% & 80.7\% & 82.5\% & 36.2\% & 68.0\% \\ \bottomrule
\end{tabular}
}
\end{small}
\vskip 0.1in
\label{tab:appendix:LLaMA-30b}
\end{table*}

\begin{table*}[t]
\centering
\caption{LLaMA-2-7B}
\vskip 0.1in
\begin{small}
\resizebox{\linewidth}{!}{
\begin{tabular}{@{}lcccccccc@{}}
\toprule
Method & WikiText2 & ARC (easy) & ARC (challenge) & LAMBADA & PIQA & BOOLQ & OpenbookQA & Avg. Accuracy \\ \midrule
FP16 & 5.48 & 76.3\% & 43.6\% & 73.9\% & 78.1\% & 77.7\% & 31.4\% & 63.5\% \\
GPTQ & 5.69 & 75.0\% & 42.2\% & 72.3\% & 77.4\% & 76.4\% & 30.0\% & 62.2\% \\
AWQ & 5.61 & 75.2\% & 43.3\% & 72.7\% & 77.6\% & 77.3\% & 31.4\% & 62.9\% \\
LLM.int4() & 5.77 & 75.1\% & 42.7\% & 71.9\% & 77.6\% & 76.2\% & 32.2\% & 62.6\% \\
OmniQuant (W6A6) & 5.87 & 67.3\% & 39.0\% & 0.0\% & 77.6\% & 69.9\% & 29.2\% & 47.2\% \\
LQER-INT (W4A8) & 5.85 & 74.7\% & 42.4\% & 71.6\% & 76.7\% & 76.1\% & 32.0\% & 62.2\% \\
LQER-MXINT (W4A6) & 5.73 & 75.1\% & 43.1\% & 73.6\% & 77.6\% & 76.2\% & 32.6\% & 63.0\% \\
LQER-MXINT (W4A8) & 5.69 & 75.3\% & 42.5\% & 73.7\% & 77.9\% & 76.3\% & 32.8\% & 63.1\% \\ \bottomrule
\end{tabular}
}
\end{small}
\vskip 0.1in
\label{tab:appendix:llama-2-7b}
\end{table*}

\begin{table*}[t]
\centering
\caption{LLaMA-2-13B}
\vskip 0.1in
\begin{small}
\resizebox{\linewidth}{!}{
\begin{tabular}{@{}lcccccccc@{}}
\toprule
Method & WikiText2 & ARC (easy) & ARC (challenge) & LAMBADA & PIQA & BOOLQ & OpenbookQA & Avg. Accuracy \\ \midrule
FP16 & 4.90 & 79.4\% & 48.3\% & 76.7\% & 79.1\% & 80.6\% & 35.0\% & 66.5\% \\
GPTQ & 5.06 & 78.6\% & 47.4\% & 76.4\% & 78.2\% & 80.8\% & 34.2\% & 65.9\% \\
AWQ & 4.98 & 78.9\% & 46.9\% & 76.2\% & 78.8\% & 80.1\% & 34.4\% & 65.9\% \\
LLM.int4() & 4.98 & 77.6\% & 47.0\% & 76.1\% & 78.9\% & 80.5\% & 34.8\% & 65.8\% \\
OmniQuant (W6A6) & 5.14 & 71.3\% & 43.8\% & 0.0\% & 78.6\% & 69.8\% & 33.0\% & 49.4\% \\
LQER-INT (W4A8) & 5.10 & 78.5\% & 47.1\% & 75.8\% & 78.6\% & 81.0\% & 34.4\% & 65.9\% \\
LQER-MXINT (W4A6) & 5.05 & 78.2\% & 46.4\% & 76.4\% & 78.3\% & 80.6\% & 34.8\% & 65.8\% \\
LQER-MXINT (W4A8) & 5.02 & 78.3\% & 47.0\% & 76.4\% & 78.8\% & 81.3\% & 34.6\% & 66.1\% \\ \bottomrule
\end{tabular}
}
\end{small}
\vskip 0.1in
\label{tab:appendix:llama-2-13b}
\end{table*}

\begin{table*}[t]
\centering
\caption{Vicuna-7B-v1.5}
\vskip 0.1in
\begin{small}
\resizebox{\linewidth}{!}{
\begin{tabular}{@{}lcccccccc@{}}
\toprule
Method & WikiText2 & ARC (easy) & ARC (challenge) & LAMBADA & PIQA & BOOLQ & OpenbookQA & Avg. Accuracy \\ \midrule
FP16 & 6.78 & 75.6\% & 43.3\% & 71.1\% & 77.3\% & 80.9\% & 33.0\% & 63.5\% \\
GPTQ & 7.07 & 75.4\% & 41.5\% & 69.4\% & 76.0\% & 81.3\% & 33.2\% & 62.8\% \\
AWQ & 7.00 & 75.0\% & 41.8\% & 70.0\% & 77.1\% & 81.5\% & 32.2\% & 62.9\% \\
LLM.int4() & 7.14 & 75.0\% & 42.6\% & 69.3\% & 76.3\% & 81.3\% & 34.2\% & 63.1\% \\
LQER-MXINT (W4A8) & 7.01 & 75.4\% & 42.2\% & 68.9\% & 77.1\% & 81.6\% & 33.0\% & 63.0\% \\ \bottomrule
\end{tabular}
}
\end{small}
\vskip 0.1in
\label{tab:appendix:vicuna-7b}
\end{table*}

\begin{table*}[t]
\centering
\caption{Vicuna-13B-v1.5}
\vskip 0.1in
\begin{small}
\resizebox{\linewidth}{!}{
\begin{tabular}{@{}lcccccccc@{}}
\toprule
Method & WikiText2 & ARC (easy) & ARC (challenge) & LAMBADA & PIQA & BOOLQ & OpenbookQA & Avg. Accuracy \\ \midrule
FP16 & 5.92 & 78.7\% & 47.8\% & 73.4\% & 78.9\% & 85.2\% & 36.8\% & 66.8\% \\
GPTQ & 6.00 & 77.9\% & 46.4\% & 72.9\% & 78.1\% & 85.0\% & 36.8\% & 66.2\% \\
AWQ & 6.03 & 78.3\% & 48.4\% & 72.9\% & 78.3\% & 84.8\% & 36.8\% & 66.6\% \\
LLM.int4() & 6.09 & 77.5\% & 47.3\% & 73.0\% & 78.3\% & 85.2\% & 36.8\% & 66.4\% \\
LQER-MXINT (W4A8) & 6.04 & 78.5\% & 46.7\% & 72.7\% & 77.7\% & 85.0\% & 36.4\% & 66.2\% \\ \bottomrule
\end{tabular}
}
\end{small}
\vskip 0.1in
\label{tab:appendix:Vicuna-13b}
\end{table*}

\begin{table*}[t]
\centering
\caption{Mistral-7B}
\vskip 0.1in
\begin{small}
\resizebox{\linewidth}{!}{
\begin{tabular}{@{}lcccccccc@{}}
\toprule
Method & WikiText2 & ARC (easy) & ARC (challenge) & LAMBADA & PIQA & BOOLQ & OpenbookQA & Avg. Accuracy \\ \midrule
FP16 & 6.47 & 82.7\% & 53.5\% & 70.7\% & 80.4\% & 86.2\% & 32.8\% & 67.7\% \\
GPTQ & 8.13 & 81.1\% & 55.8\% & 72.2\% & 80.9\% & 86.7\% & 36.0\% & 68.8\% \\
AWQ & 6.64 & 81.9\% & 53.8\% & 71.8\% & 80.7\% & 86.2\% & 37.4\% & 68.6\% \\
LLM.int4() & 6.66 & 81.2\% & 53.2\% & 70.6\% & 81.2\% & 86.4\% & 34.6\% & 67.9\% \\
LQER-MXINT (W4A8) & 6.71 & 81.7\% & 53.8\% & 71.2\% & 81.0\% & 86.5\% & 34.8\% & 68.2\% \\ \bottomrule
\end{tabular}
}
\end{small}
\vskip 0.1in
\label{tab:appendix:mistral-7b}
\end{table*}


\end{document}